\documentclass[10pt,twocolumn,letterpaper]{article}

\usepackage[pagenumbers]{cvpr} 

\definecolor{cvprblue}{rgb}{0.21,0.49,0.74}
\usepackage[pagebackref,breaklinks,colorlinks,allcolors=cvprblue]{hyperref}
\usepackage{multirow}
\usepackage{algorithm}
\usepackage{algpseudocode}
\usepackage{listings}
\usepackage{xcolor}
\usepackage{tcolorbox}

\def\paperID{***} 
\def\confName{CVPR}
\def\confYear{2026}

\title{L1 Sample Flow for Efficient Visuomotor Learning}

\author{
    Weixi Song$^{1,2,3}$ \quad Zhetao Chen$^{1,2}$ \quad Tao Xu$^{2}$ \quad Xianchao Zeng$^2$ \quad Xinyu Zhou$^2$ \quad Lixin Yang$^{2,4}$ \\ Donglin Wang$^{3\dagger}$ \qquad Cewu Lu$^{2,4}$ \qquad Yong-Lu Li$^{2,4 \dagger}$ \\
    $^1$Zhejiang University \qquad $^2$Shanghai Innovation Institute \qquad 
    $^3$Westlake University\\ 
    $^4$Shanghai Jiao Tong University
    \\
    {\tt\small songweixi@zju.edu.cn \qquad yonglu\_li@sjtu.edu.cn}
}

\begin{document}
\maketitle

\begingroup
\renewcommand\thefootnote{}\footnotetext{
$^\dagger$Corresponding author. 
}
\addtocounter{footnote}{-1}
\endgroup

\begin{abstract}
Denoising-based models, such as diffusion and flow matching,  have been a critical component of robotic manipulation for their strong distribution-fitting and scaling capacity. Concurrently, several works have demonstrated that simple learning objectives, such as L1 regression, can achieve performance comparable to denoising-based methods on certain tasks, while offering faster convergence and inference. 
In this paper, we focus on how to combine the advantages of these two paradigms: retaining the ability of denoising models to capture multi-modal distributions and avoid mode collapse while achieving the efficiency of the L1 regression objective. 
To achieve this vision, we reformulate the original v-prediction flow matching and transform it into sample-prediction with the L1 training objective. We empirically show that the multi-modality can be expressed via a single ODE step. Thus, we propose \textbf{L1 Flow}, a two-step sampling schedule that generates a suboptimal action sequence via a single integration step and then reconstructs the precise action sequence through a single prediction. The proposed method largely retains the advantages of flow matching while reducing the iterative neural function evaluations to merely two and mitigating the potential performance degradation associated with direct sample regression. We evaluate our method with varying baselines and benchmarks, including 8 tasks in MimicGen, 5 tasks in RoboMimic \& PushT Bench, and one task in the real-world scenario. The results show the advantages of the proposed method with regard to training efficiency, inference speed, and overall performance.  \href{https://song-wx.github.io/l1flow.github.io/}{Project Website.}
\end{abstract}    
\section{Introduction}
\label{sec:intro}
\begin{figure}[t]
  \centering
  \includegraphics[width=\linewidth]{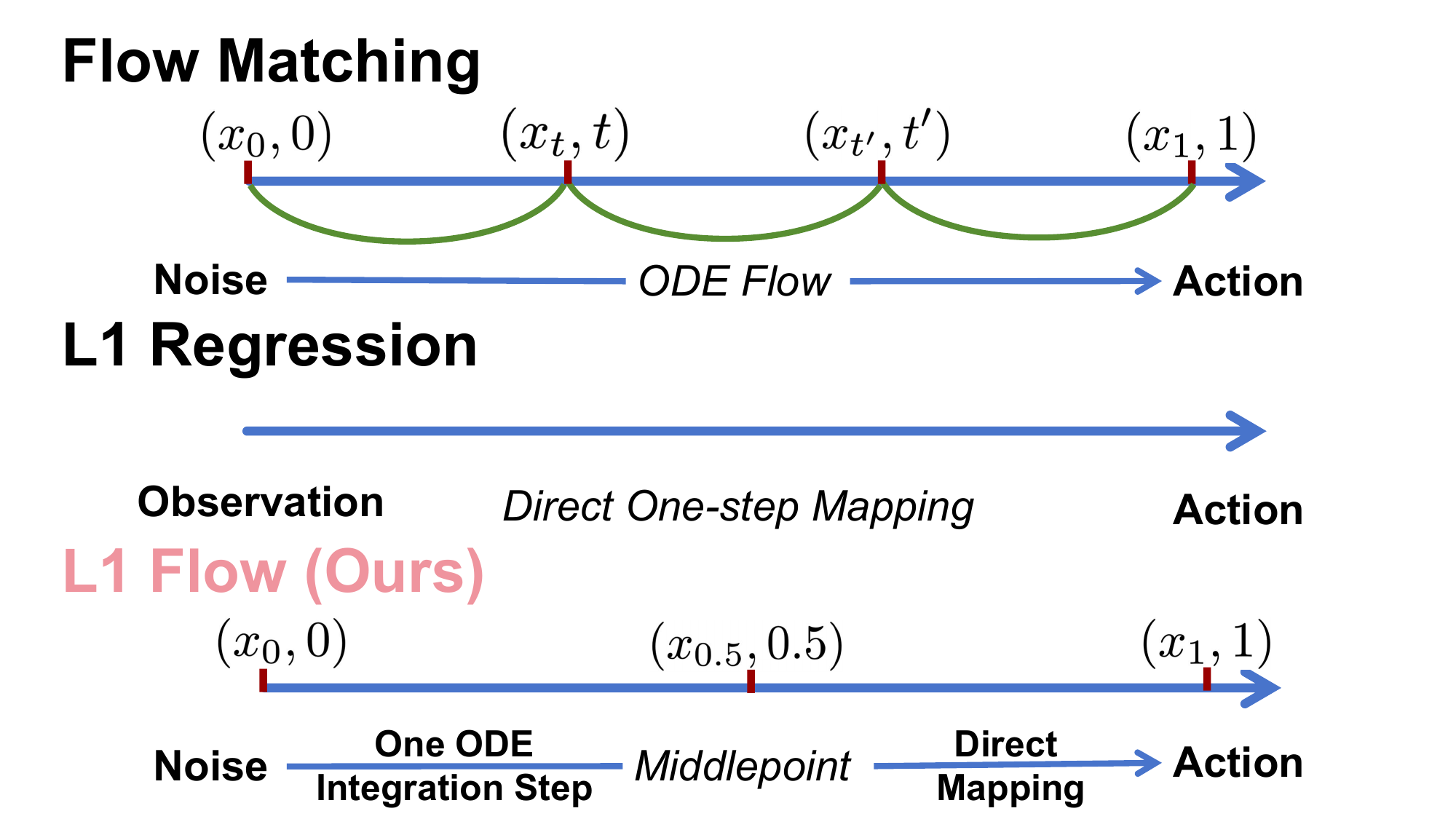}
    \caption{\textbf{Overview of the proposed method.} \textbf{L1 Flow} employs a 2-step denoising paradigm which combines the efficiency of L1 regression and the strong distribution-modeling capacity of standard flow matching. Compared with the iterative denoising process of the standard flow matching and the direct mapping of L1 regression, L1 Flow decouples the modeling of multi-modal distribution and the reconstruction of the precise actions. Starting from a random noise, L1 Flow performs one integration step towards the middle timestep and predicts the precise action \(x_1\) from the coarse action \(x_{0.5}\), which are based on the reformulated sample-prediction type flow matching. }
  \label{fig:sum}
\end{figure}

Learning from demonstrations to map the observations to actions formulates the basic paradigm of visuomotor policy learning, casting the cloning of expert behavior as supervised learning. Denoising-based methods, like diffusion and flow matching, play a significant role in robot manipulation imitation learning due to their powerful distribution modeling capabilities, with applications ranging from lightweight visuomotor policies\citep{chi2023diffusion,wang2024equivariant,wang2024sparse,ze20243d, zhang2025flowpolicy} to large scale Visual-Language-Action models\citep{liu2024rdt,wen2025dexvla,wen2025tinyvla,wen2025diffusionvla,black2024pi0visionlanguageactionflowmodel,rdt2,intelligence2025pi05visionlanguageactionmodelopenworld,shukor2025smolvla}. However, the multi-timestep noise prediction in training and the iterative denoising process during inference lead to slow training convergence and high inference latency, respectively, which pose a challenge for their practical application in robotics tasks. 
To accelerate the inference process, \citet{wang2024one} distills the diffusion policy into a one-step action generator by minimizing the KL divergence along the diffusion chain, which introduces additional post-training cost after the initial training of the diffusion policy. 
Non-denoising approaches\cite{mandlekar2021matters,zhao2023learning,florence2022implicit, shafiullah2022behavior} have long been proposed for visuomotor policy learning, which map observations directly to actions through simple regression, allowing for efficient policy training. These methods were once considered inferior to denoising-based approaches due to their limited ability to capture complex, especially multi-modal, action distributions. However, with appropriate architectural modifications\cite{su2025dense,gong2025carp}, the direct action regression objective can achieve performance comparable to denoising-based methods. Recently, \citet{kim2025fine} demonstrates that simple learning objectives, such as L1 regression, can achieve performance comparable to diffusion-based methods on certain tasks, but \citet{kim2025fine} also points out that it may struggle to capture the multi-modality in human demonstrations. 

The deterministic sampling, \eg, DDIM\cite{song2020denoising} and flow matching\cite{lipman2022flow}, are commonly used to accelerate the generation of action sequences for real-time inference, which means the model’s ability to capture multi-modal behaviors is derived solely from the stochastic sampling of the initial noise. 
Therefore, we are motivated to model the multi-modality via the minimal integration step and reconstruct the precise action sequences through the direct L1 regression conditioned on the integral result, which combines the strong distributional expressiveness of denoising models with the efficient training and inference features of L1 regression. 
To achieve this, we reformulate the original v-prediction form of flow matching into an equivalent sample-prediction formulation, enabling the direct regression on the target samples. Based on this sample-prediction variant, we introduce a two-step sampling procedure.
In the first step, a noisy sample drawn at $t=0$ is transformed into a coarse sample by converting the model’s prediction into the corresponding velocity of the original flow matching and performing a single-step ODE integration. Empirically, we find that this single integration step is sufficient to capture the modality of the target distribution as shown in Figure \ref{fig:mode_vis}. In the second step, the model directly predicts the final clean sample conditioned on the coarse sample, resulting in efficient reconstructions that preserve the multimodality of the data distribution.

We first compare the proposed two-step sampling strategy with 2 standard denoising-based methods \(\ie\) DDPM (100 steps) and Flow Matching (10 steps), and L1 regression in 8 tasks of MimicGen Benchmark. The results (Figure \ref{fig:mimicgen}) and Table \ref{exp:mimicgen_all}) exhibit the better training efficiency and comparable overall performance of L1 Flow, though it is sampled with much fewer times. Then, as an efficient paradigm aiming at speeding up, L1 Flow is compared to distillation-based methods \(\ie\) Consistency Policy (CP)\cite{prasad2024consistency} and OneDP\cite{wang2024one} in total 5 tasks of Robomimic and PushT Bench. In contrast to the pretraining-distillation training of distillation-based methods, we apply end-to-end direct training, achieving faster training while outperforming in most tasks. Further, we establish a 2-stage task in real-world scenarios, mainly focusing on multi-modality and prediction precision. The results show the advantage of L1 Flow in overall performance and about 10\(\times\) to 70\(\times\) faster inference speed than the common diffusion policy.

In conclusion, the main contributions are as follows:
\begin{itemize}
    \item We reformulate the origin velocity prediction flow matching to sample prediction and empirically show that the multi-modality can be expressed via a single integration step.
    \item Based on the reformulated sample-prediction flow matching, we propose an efficient two-step denoising strategy to reconstruct the action sequence via one-step integration and direct sample prediction and validate its effectiveness in both simulation and real-world scenarios.
\end{itemize}

\section{Related Work}

\subsection*{Denoising-based Models in Robotics Manipulation}
Recent advances have shown that denoising models can serve as powerful policies for robot manipulation, including imitation learning\cite{rdt2,liu2024rdt,black2024pi0visionlanguageactionflowmodel,chi2023diffusion,wen2025diffusionvla,wang2024equivariant,ze20243d} and reinforcement learning\cite{sun2025scorebased, park2025flow, ren2025diffusion,wang2023diffusion}, due to their strong distribution mapping capacity. Diffusion Policy\cite{chi2023diffusion} formulates visuomotor control as conditional denoising of action sequences, enabling robust and multi-modal policy learning across diverse tasks, with extensions like DP3\cite{ze20243d} incorporating the 3D visual representations into diffusion policies and EquiDP\cite{wang2024equivariant} considering the domain symmetries in manipulation. Subsequent works~\cite{liu2024rdt,wen2025tinyvla,wen2025diffusionvla,wen2025dexvla} scale this paradigm to large pre-trained models, achieving strong generalization across varied robot embodiment and deployment scenes. More recently, flow-matching\cite{lipman2022flow, liu2022flow} approaches have been widely applied to large robotics foundation models\cite{black2024pi0visionlanguageactionflowmodel,intelligence2025pi05visionlanguageactionmodelopenworld,rdt2,shukor2025smolvla} for faster, deterministic sampling, but they still need about 10 NFE. 

To ease the time-consuming iterative denoising process, distillation-based methods \citep{wang2024one, prasad2024consistency} accelerate the sampling by training one-/ few-step student models with pre-trained teacher models as prior. Consistency Policy(CP)\cite{prasad2024consistency} adapts the Consistency Model frameworks\cite{kim2024consistency,song2024improved,Song2023ConsistencyM} and achieves faster inference speed while maintaining comparable performances against the baseline. To address slow convergence and occasional performance degradation in CP, OneDP\cite{wang2024one} distills the diffusion policy into a one-step action generator by minimizing the KL divergence along the diffusion chain. Distillation-based methods introduces additional post-training cost after the initial training. For end-to-end training, DM1\cite{zou2025dm1} applies MeanFlow\cite{geng2025mean}, a recent one-step generation framework, to achieve one-step generation. However, it also brings an additional training burden in the computation of Jacobian–Vector Product (JVP) bringing more 15\% overhead. 
\subsection*{Non-Denoising Models in Robotics Manipulation}
There has long been a series of non-denoising methods for learning visuomotor policy\cite{mandlekar2021matters,zhao2023learning,florence2022implicit, shafiullah2022behavior}, which directly map observations to actions via simple regression, achieving efficient policy training. For a period of time, such alternatives were considered inferior to denoising-based methods for their poor capacity to model complex, especially, multi-modal distributions. Several architectural modifications\cite{su2025dense,gong2025carp} \(\eg\) auto-regressive multi-scale prediction, together with the direct action regression objective, can achieve competitive performance against denoising-based approaches. However, OpenVLA-OFT\cite{kim2025fine} points out that using L1 regression with parallel decoding can achieve performance comparable to diffusion-based methods on certain tasks, and VLA-Adapter\cite{wang2025vla} also achieves state-of-the-art results on various benchmarks with L1 training objective. But the fact is that L1 regression is hard to capture the multi-modality in human demonstrations and is easy to encountering mode collapse when trained with similar inputs but largely different outputs. This work aims to maintain the training and inference efficiency of simple regression while including the strong distribution-mapping capacity of denoising-based methods. In the following section, we establish a sample prediction type of flow matching and introduce L1 Flow to bridge simple regression and multi-modality modeling via two-step denoising. 
\begin{figure}[t]
  \centering
  \includegraphics[width=\linewidth]{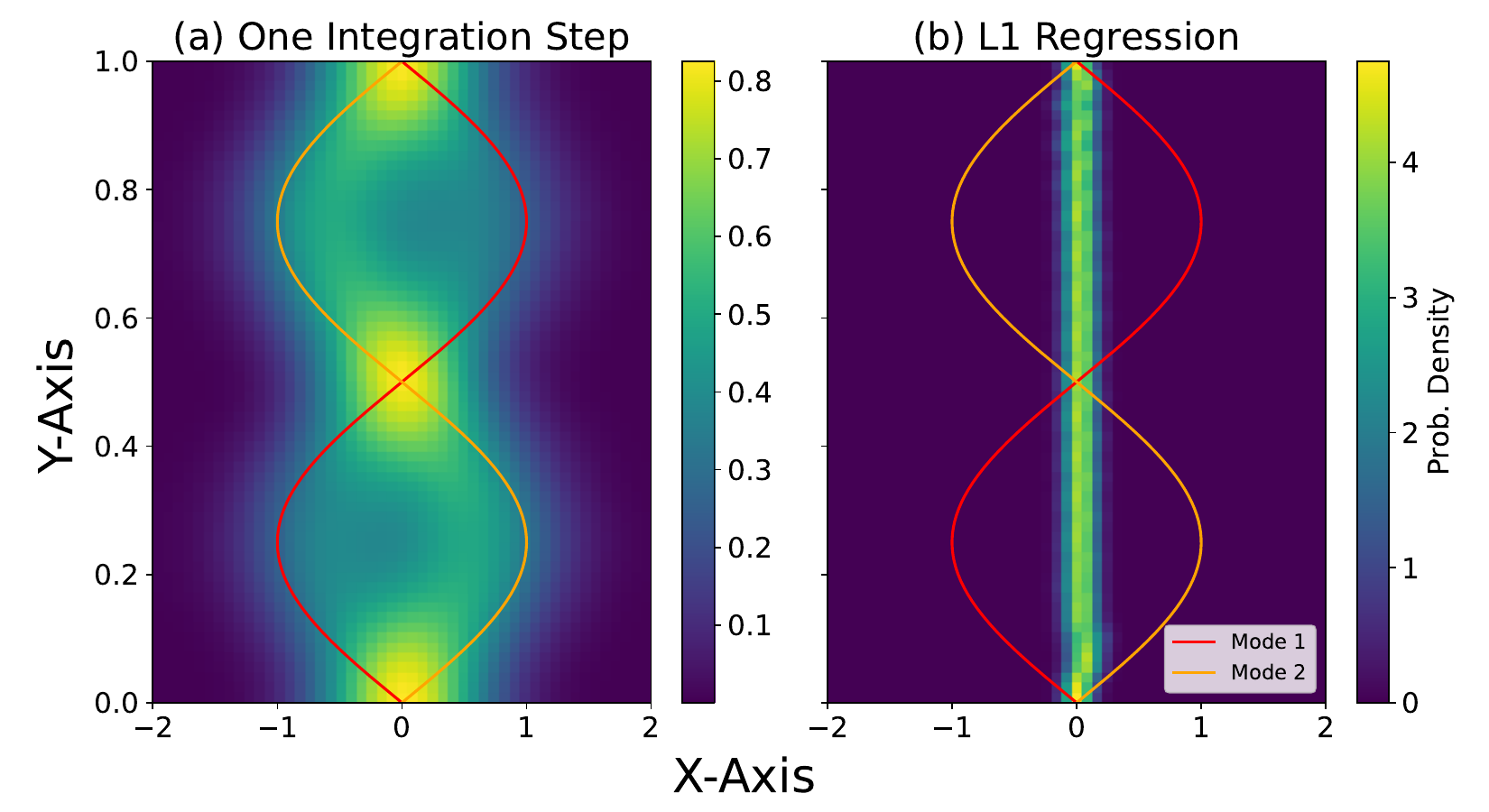}
  \caption{\textbf{Visualization of the sample distribution.} Apply the proposed one-step integration to model two sine curves with different phases and compare with L1 Regression. \textbf{(a)} One-step integration effectively captures the multi-modality. \textbf{(b)} The direct regression exhibits the average of two modes, the so-called mode collapse.}
  \label{fig:mode_vis}
\end{figure}

\begin{figure}[t]
  \centering
  \includegraphics[width=0.8\linewidth]{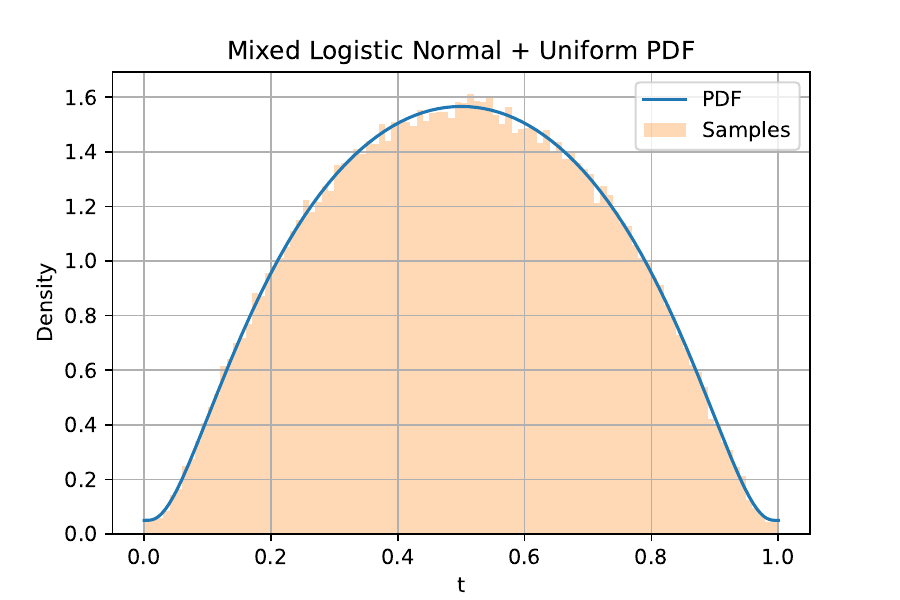}
  \caption{\textbf{PDF of the mixed distribution of Logistic Normal and Uniform distribution.} The Logistic Normal distribution emphasizes sampling around intermediate timesteps, while we additionally incorporate a low-level uniform distribution to ensure that the probabilities at the boundary timesteps remain non-zero.}
  \label{fig:mix-ln-dist}
\end{figure}
\section{L1 Sample Flow}
\subsection{Preliminary}
Flow matching \cite{lipman2022flow} formulates generative modeling as learning a continuous-time deterministic flow that transports a simple base distribution $p(x_0)$, \(\eg\) a Gaussian distribution, toward a complex target distribution $p(x_1)$. The goal of flow matching is to learn a velocity field 
$v_\theta(x_t,t)$ such that the solution of the following ordinary differential equation (ODE) satisfies $x_1\sim p(x_1)$:
\begin{equation}
    \frac{dx_t}{dt}=v_\theta(x_t,t).
\end{equation}
A common flow choice is the linear interpolation path~\cite{liu2022flow}:
\begin{equation}
    x_t = (1-t)x_0+tx_1, t\in[0,1],
\end{equation}
which yields the ground-truth velocity:
\begin{equation}
    v_\theta(x_t,t)=\frac{dx_t}{dt}=x_1-x_0.
\end{equation}
The training objective minimizes the discrepancy between the predicted flow and the target flow:
\begin{equation}
   \mathcal{L}(\theta)=\mathbb{E}_{x_0,x_1,t}||\textcolor{red}{f_\theta(x_t,t)}-(x_1-x_0)||^2.
\end{equation}
Samples are generated by integrating the learned flow field from 
$t=0$ to $t=1$, starting from random noise:
\begin{equation}
    x_1=x_0+\int_0^1f_\theta(x_t,t)dt.
\end{equation}

\subsection{Methodology}
\label{sec:methodology}
We consider a variant of Flow Matching where the model directly predicts the terminal sample rather than the instantaneous velocity.
Instead of learning the instantaneous velocity field \(x_1-x_0\),
the model $\textcolor{red}{f_\theta(x_t,t)}$ predicts the corresponding terminal sample $x_1$ conditioned on the intermediate state $x_t$ and time $t$:
\begin{equation}
    \textcolor{red}{f_\theta(x_t,t)}=\frac{dx_t}{dt}=x_1-x_0\Rightarrow x_1.
\end{equation}
The instantaneous velocity can then be implicitly recovered as: 
\begin{equation}
\begin{aligned}
      v=x_1-x_0&=\frac{(1-t)x_1-(1-t)x_0}{1-t}\\
    &=\frac{x_1-((1-t)x_0+tx_1)}{1-t}\\
    &=\frac{\textcolor{red}{f_\theta(x_t,t)}-x_t}{1-t}.  
\end{aligned}
\end{equation}
This yields a sample-prediction flow whose dynamics are defined by the ODE:
\begin{equation}
    \frac{dx}{dt}=\frac{\textcolor{red}{f_\theta(x_t,t)}-x_t}{1-t}.
\end{equation}
At training time, the model is optimized to minimize the discrepancy between the predicted terminal sample $x_{pred}$ and the true target sample $x_1$. Similar to \cite{kim2025fine,wang2025vla}, we also employ L1 loss to supervise the target samples:
\begin{equation}
\mathcal{L}=\mathbb{E}_{x_0\sim\mathcal{N}(0,1),x_1\sim data,t}\ ||\textcolor{red}{f_\theta(x_t,t)}-x_1||_1.
\label{eq:x_loss_L1}
\end{equation}
Empirically, we observe that under this setting, the L1 loss performs better than the MSE loss, which is commonly used in standard flow matching to supervise the velocity field. A further comparison is included in the ablation study in Section \ref{sec:ablation}. 

During inference, we introduce a two-step denoising schedule that combines continuous flow integration with direct sample prediction.
Starting from a pure noise initialization $x_0\sim \mathcal{N}(0,1)$, we first integrate the learned flow field from $t=0$ to an intermediate time $t=0.5$:
\begin{equation}
    x_{0.5}=x_0+\frac{f_\theta(x_0,0)-x_0}{2}.
\end{equation}
This partial integration allows the model to evolve the noisy sample toward an intermediate representation $x_{0.5}$ that captures the coarse structural and modality information about the target distribution.
At the midpoint, instead of continuing the ODE integration until
$t=1$, we directly invoke the model's sample prediction capability to obtain the final output:
\begin{equation}
    x_{1} = \textcolor{red}{f_\theta(x_{0.5}, 0.5)}.
\end{equation}
We leverage the multi-modal capture ability of flow matching to predict a coarse sample at mid-timestep while utilizing the efficiency of simple sample prediction that avoids time-consuming integration. Note that our two-step schedule is mathematically equivalent to the standard 2-step integration. The numerical errors in the calculation of the velocity bring the performance gap(see ablation study in Section \ref{sec:inference_ablation}). Therefore, we apply the current 2-step version. We apply the proposed sample-prediction flow matching to model two sine curves with different phases (simulating different trajectories of human demonstrations under the same observations). Samples are generated via a single-step integration, and the resulting sample distribution is visualized as the heatmap shown in Figure \ref{fig:mode_vis}(a). For comparison, we also perform the direct L1 regression on the curves, where Gaussian noise is used as input to enable one-to-many mapping in the regression. The results demonstrate that our one-step integration effectively captures the multi-modality of the distribution, while the direct regression exhibits the average of two modes, the so-called mode collapse.

With coarse samples, the subsequent prediction is to accurately reconstruct samples from different modalities. To further enhance the model’s capability of predicting samples at the intermediate timestep, we sample the timestep $t$ from the mixed distribution of logistic-normal distribution and the uniform one:
\begin{equation}
\label{eq:mlog}
    t =
\begin{cases}
\frac{1}{1+e^{-x}}, x\sim \mathcal{N}(0,1), &p=1-\alpha, \\
x,x\sim U(0,1), & \text{otherwise}.
\end{cases}
\end{equation}
While increasing the sampling probability at intermediate time steps, the sampling probabilities at the boundary timesteps are maintained at appropriately low levels, as shown in the visualization in Figure \ref{fig:mix-ln-dist}.
As a result, the model develops stronger representational capacity around intermediate timesteps, which directly benefits the midpoint-prediction sampling procedure described above. We summarize the training and inference outlines of our proposed method in Algorithm.\ref{alg:training} and \ref{alg:infer}.

\begin{algorithm}[t]
\caption{L1 Sample Flow Training}
\label{alg:training}
\begin{algorithmic}[1]
\Require Dataset of pairs of observations and trajectories $\mathcal{D} = \{ (o,x_1)\}$, model $f_\theta$
\While{not converged}
    \State Sample a batch of pairs of observations and trajectories $(o,x_1) \sim \mathcal{D}$
    \State Sample a batch of noise $x_0 \sim \mathcal{N}(0,1)$
    \State Sample $t$ by Equation(\ref{eq:mlog}) 
    \State Compute interpolated state: $x_t = (1 - t)x_0 + t x_1$
    \State Predict terminal sample: $\hat{x}_1 = f_\theta(x_t,t,o)$
    \State Compute L1 loss: $\mathcal{L} = \|\hat{x}_1 - x_1\|_1$
    \State Update $\theta \gets \theta - \eta \nabla_\theta \mathcal{L}$
\EndWhile
\end{algorithmic}
\end{algorithm}

\begin{algorithm}[t]
\caption{L1 Sample Flow Inference}
\label{alg:infer}
\begin{algorithmic}[1]
\Require model $f_\theta$, observation $o$
\State Sample noise $x_0 \sim \mathcal{N}(0,1)$
\State $x_{pred}=f_\theta(x_0,0,o)$
\State $x_{0.5}=x_0+\frac{1}{2}(f_\theta(x_0,0)-x_0)$ \Comment{One-step Integration}
\State $x_1=f_\theta(x_{0.5},0.5)$ \Comment{Direct Sample Prediction}
\end{algorithmic}
\end{algorithm}

\section{Experiment}
\begin{figure*}[t]
  \centering
  \begin{subfigure}{0.24\linewidth}
    \includegraphics[width=\linewidth]{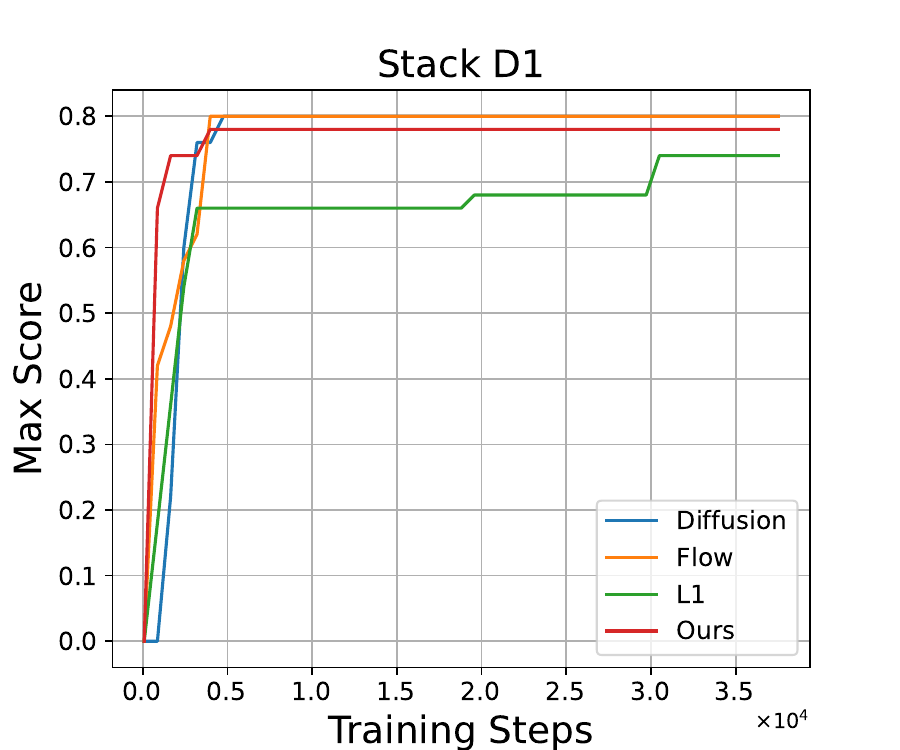}
    \label{fig:exp-stack}
  \end{subfigure}
  \begin{subfigure}{0.24\linewidth}
    \includegraphics[width=\linewidth]{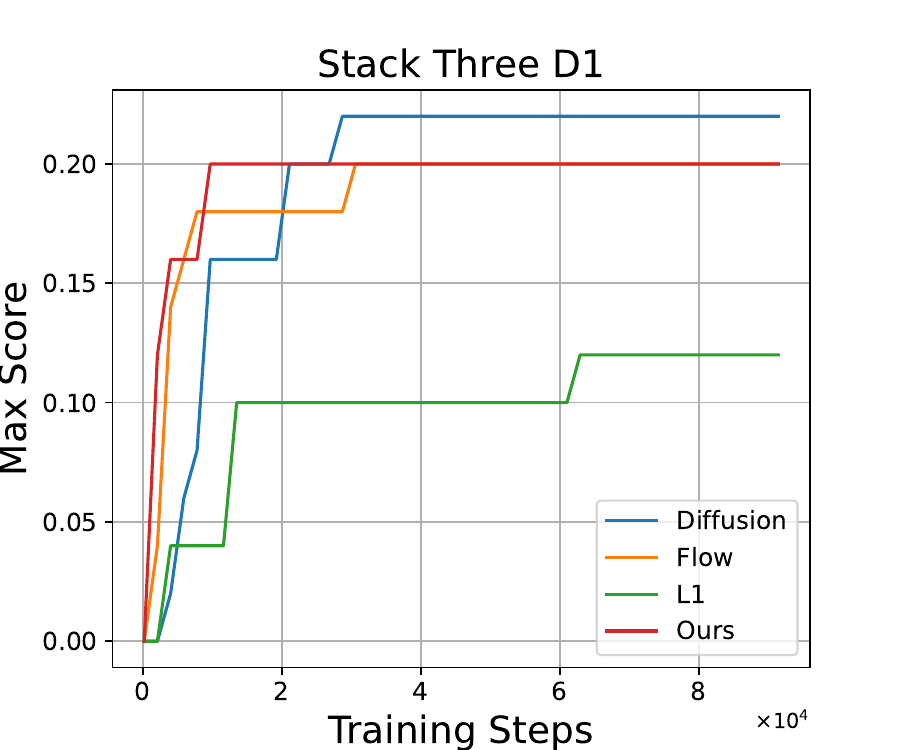}
    \label{fig:exp-stack_three}
  \end{subfigure}
  \begin{subfigure}{0.24\linewidth}
    \includegraphics[width=\linewidth]{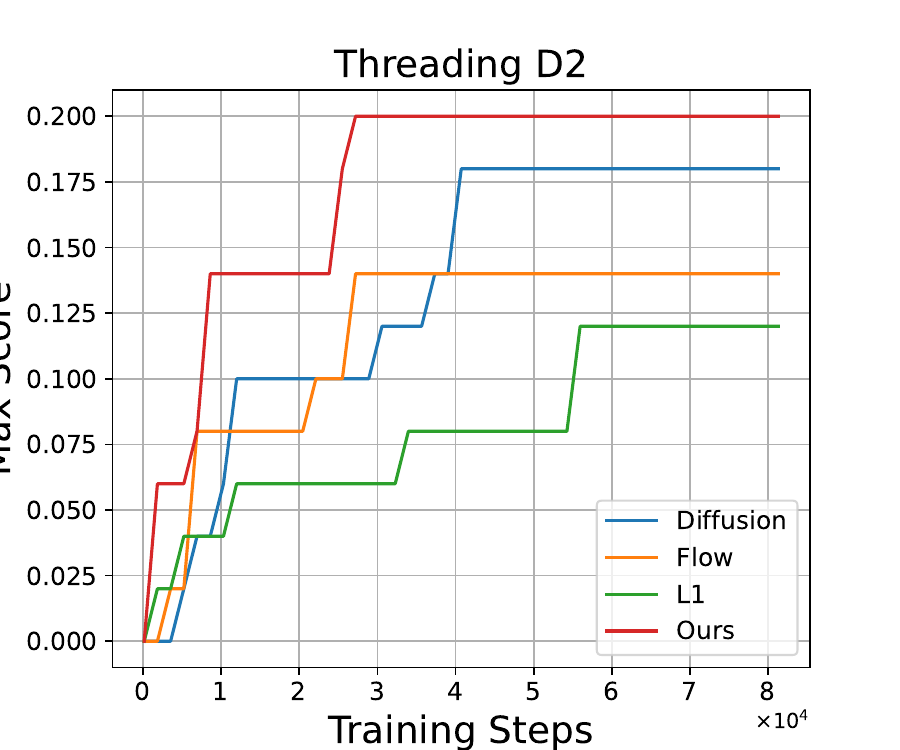}
    \label{fig:exp-threading}
  \end{subfigure}
  \begin{subfigure}{0.24\linewidth}
    \includegraphics[width=\linewidth]{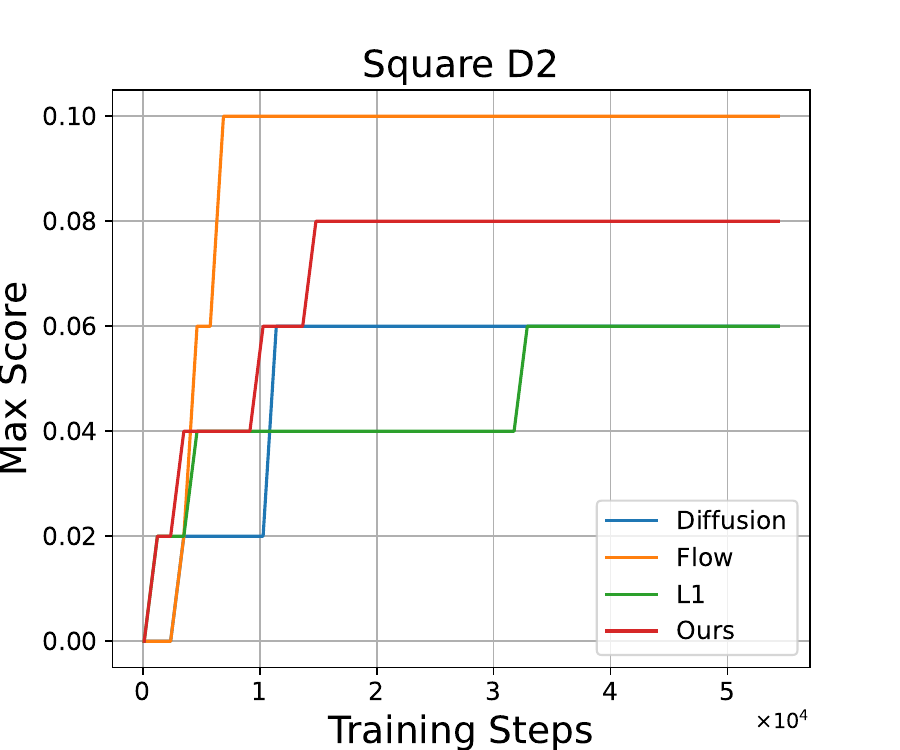}
    \label{fig:exp-square}
  \end{subfigure}
  \begin{subfigure}{0.24\linewidth}
    \includegraphics[width=\linewidth]{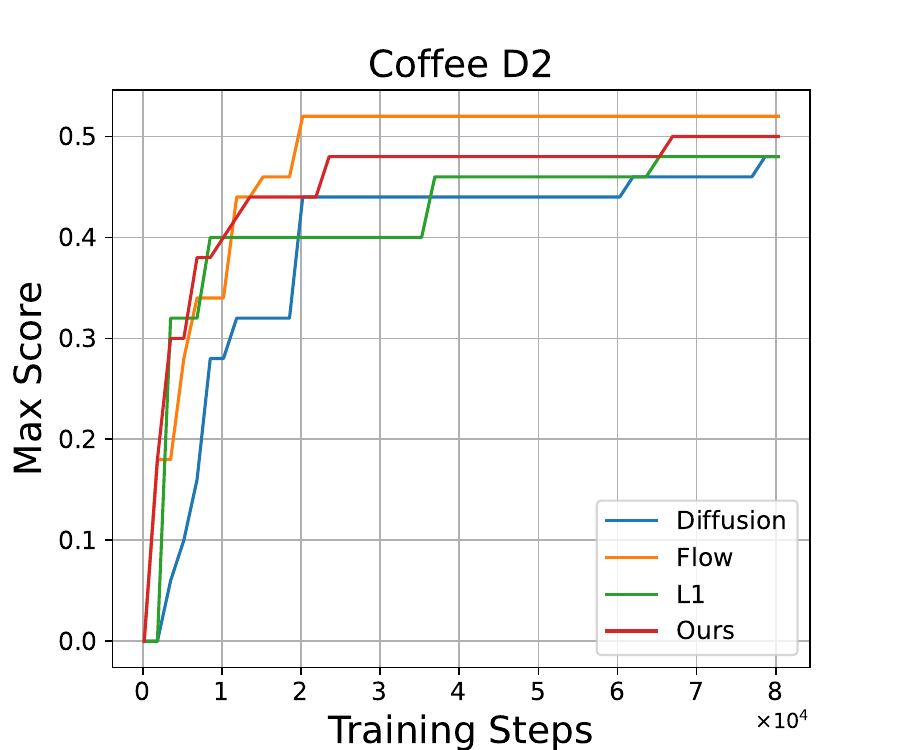}
    \label{fig:exp-coffee}
  \end{subfigure}
  \begin{subfigure}{0.24\linewidth}
    \includegraphics[width=\linewidth]{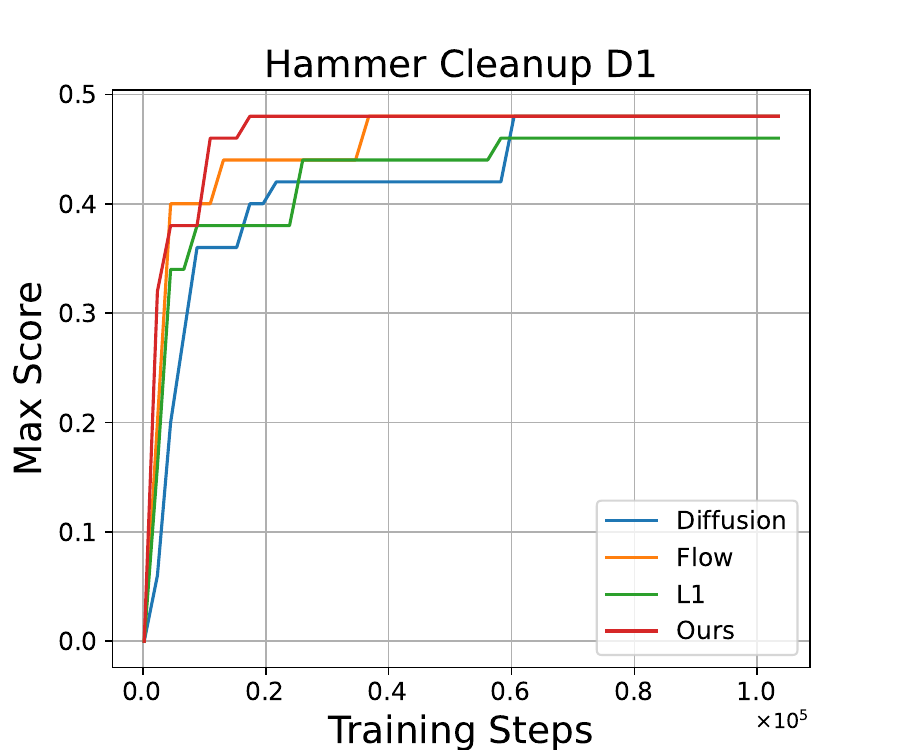}
    \label{fig:exp-hammer}
  \end{subfigure}
  \begin{subfigure}{0.24\linewidth}
    \includegraphics[width=\linewidth]{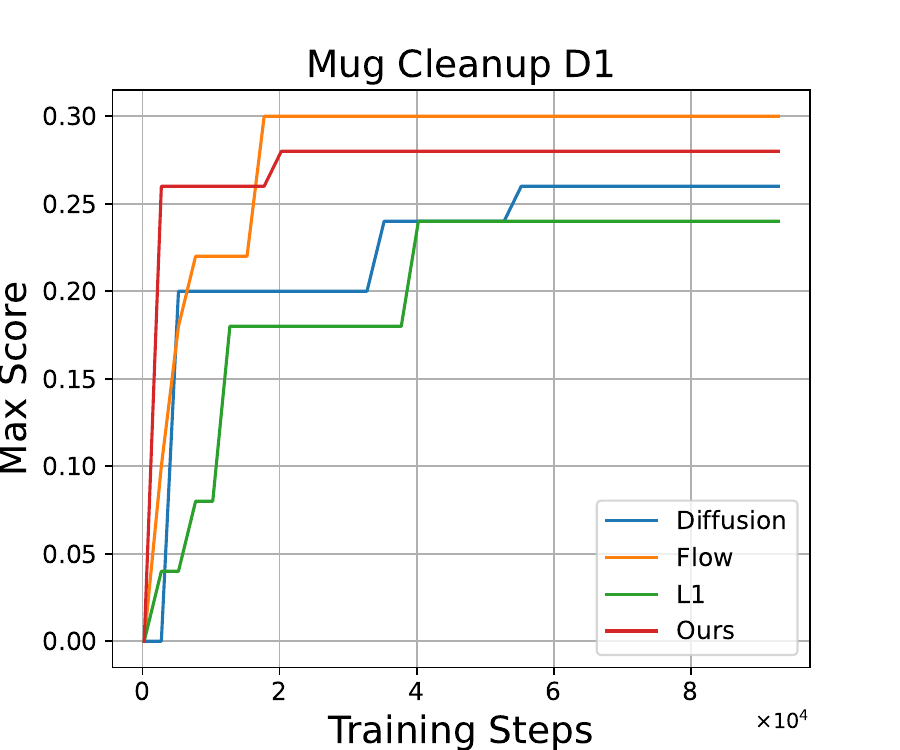}
    \label{fig:exp-mug}
  \end{subfigure}
  \begin{subfigure}{0.24\linewidth}
    \includegraphics[width=\linewidth]{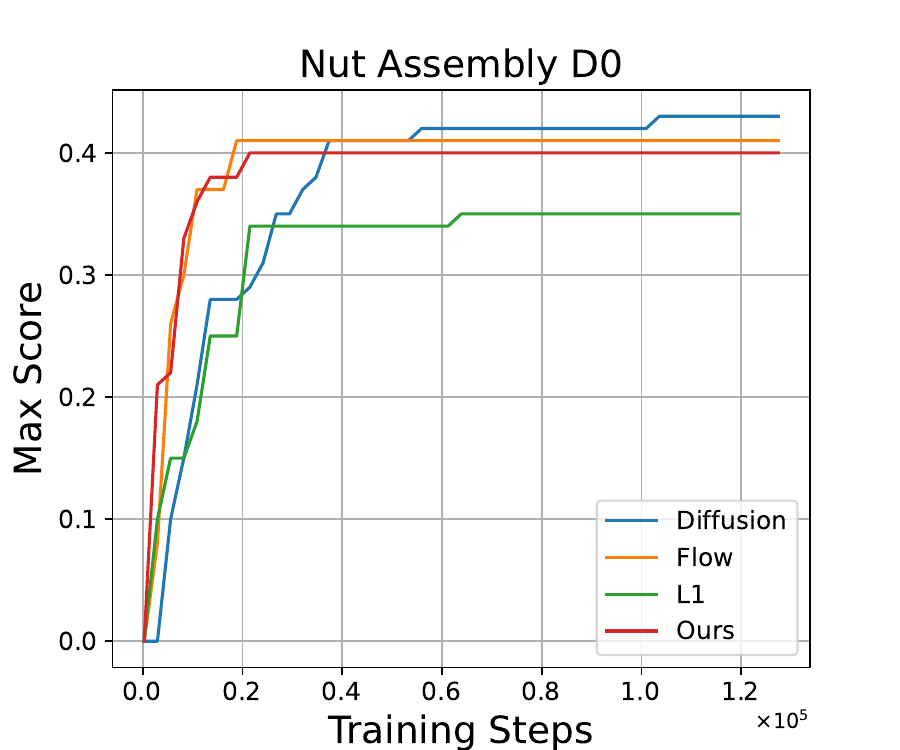}
    \label{fig:exp-nut}
  \end{subfigure}

  \caption{\textbf{The trend of the maximum success rate throughout the training.} We experiment with our method with baselines in the 8 tasks of MimicGen and report the maximum success rate among 50 evaluations throughout the training. The overall results demonstrate that our method achieves performance comparable to or even surpassing the baselines with only two neural function evaluations (NFE), while also exhibiting higher training efficiency by reaching performance saturation in success rate with fewer training steps. L1 Flow largely retains the advantages of flow matching with less inference budget while mitigating the potential performance degradation associated with direct L1 regression.}
  \label{fig:mimicgen}
\end{figure*}

In this section, we comprehensively evaluate the proposed method with varying baselines in both simulation and real-world settings. The analysis focuses mainly on the following aspects:

\begin{itemize}
    \item Compared with standard flow matching, diffusion, and L1 regression, does our method outperform in terms of training efficiency, inference efficiency, and performance?
    \item Compared with related accelerated denoising-based methods \eg, One-DP\cite{wang2024one} and Consistency Policy\cite{prasad2024consistency}, does our approach demonstrate advantages?
    \item When applied to real-world scenarios, how does our approach remain effective and benefit from faster inference?
    \item What are the contributions of each component in our method, and how does our approach connect to standard flow matching?
\end{itemize}

\begin{table*}[t]
    \caption{\textbf{Performance comparison with fast denoising baselines on RoboMimic.}  We compare our method \textbf{L1 Flow (Ours)} against representative fast denoising baselines, including full Diffusion Policies (DP) trained under DDPM, DDIM, and EDM schedulings, as well as distillation-based approaches—Consistency Policy (CP) and OneDP. We report the mean and standard deviation of success rates over five independent training runs, each evaluated under 100 randomized environment initializations (500 trials in total). Results marked with * are taken from \cite{wang2024one} and the best results are marked in \textbf{bold}. As shown in the table, \textbf{L1 Flow} achieves the best average success rates, converging about \textbf{5-7\(\times\) faster} than the baselines while requiring only two inference steps (NFE = 2).}
    \label{exp:robomimic}
    \begin{center}
    \begin{small}
    \begin{sc}
    \renewcommand{\tabcolsep}{4pt}
    \begin{tabular}{l|cc|ccccc|c}
        \toprule
         Methods & Epochs & NFE & PushT & Square-mh & Square-ph & ToolHang-ph & Transport-ph & Avg. \\
         \midrule
         DP (DDPM)* & 1000 & 100 & 0.863$\pm$0.040 & 0.846$\pm$0.023 & 0.926$\pm$0.023 & 0.822$\pm$0.016 & 0.896$\pm$0.032 & 0.871 \\
         \midrule
         \multirow{2}{*}{DP (DDIM)*} & 1000 & 10 & 0.823$\pm$0.023 & 0.850$\pm$0.013 & 0.918$\pm$0.009 & 0.828$\pm$0.016 & 0.908$\pm$0.011 & 0.865 \\
         & 1000 & 1 & 0.000$\pm$0.000 & 0.000$\pm$0.000 & 0.000$\pm$0.000 & 0.000$\pm$0.000 & 0.000$\pm$0.000 & 0.000 \\
         \midrule
         \multirow{3}{*}{DP (EDM)*} & 1000 & 35 & 0.861$\pm$0.030 & 0.810$\pm$0.026 & 0.898$\pm$0.033 & 0.828$\pm$0.019 & 0.890$\pm$0.012 & 0.857 \\
         & 1000 & 19 & 0.851$\pm$0.012 & 0.828$\pm$0.015 & 0.880$\pm$0.014 & 0.794$\pm$0.012 & 0.860$\pm$0.013 & 0.843 \\
         & 1000 & 1 & 0.000$\pm$0.000 & 0.000$\pm$0.000 & 0.000$\pm$0.000 & 0.000$\pm$0.000 & 0.000$\pm$0.000 & 0.000 \\
         \midrule
         \multirow{2}{*}{CP (EDM)*} & 1450 & 3 & 0.839$\pm$0.037 & 0.710$\pm$0.018 & 0.874$\pm$0.022 & 0.626$\pm$0.041 & 0.848$\pm$0.028 & 0.779 \\
         & 1450 & 1 & 0.828$\pm$0.055 & 0.646$\pm$0.047 & 0.776$\pm$0.055 & 0.650$\pm$0.046 & 0.754$\pm$0.120 & 0.731 \\
         \midrule
         OneDP-D (EDM)* & 1020 & 1 & 0.829$\pm$0.052 & 0.776$\pm$0.023 & 0.902$\pm$0.040 & 0.762$\pm$0.056 & 0.898$\pm$0.019 & 0.833 \\
         OneDP-S (EDM)* & 1020 & 1 & 0.841$\pm$0.042 & 0.774$\pm$0.003 & 0.910$\pm$0.041 & 0.824$\pm$0.039 & 0.910$\pm$0.027 & 0.852 \\
         \midrule
         OneDP-D (DDPM)* & 1020 & 1 & 0.802$\pm$0.057 & 0.846$\pm$0.028 & 0.926$\pm$0.011 & 0.808$\pm$0.046 & 0.896$\pm$0.013 & 0.856 \\
         OneDP-S (DDPM)* & 1020 & 1 & 0.816$\pm$0.058 & 0.864$\pm$0.042 & 0.926$\pm$0.018 & \textbf{0.850$\pm$0.033} & 0.914$\pm$0.021 & 0.874 \\
         \midrule 
         L1 Flow (Ours) & 200 & 2 & \textbf{0.869$\pm$0.007} & \textbf{0.868$\pm$0.018} & \textbf{0.948$\pm$0.013} & 0.788$\pm$0.029 & \textbf{0.932$\pm$0.013} & \textbf{0.881} \\
        \bottomrule
    \end{tabular}
    \end{sc}
    \end{small}
    \end{center}
\end{table*}

\subsection{Comparison with Standard Methods}
To verify the effectiveness of the proposed method, we first evaluate our method with several standard baselines, including 2 denoising-based methods and direct L1 regression: 
\textbf{Diffusion}: We follow the initial setting of Diffusion Policy\cite{chi2023diffusion} and apply DDPM sampling with 100 denoising steps. 
\textbf{Flow Matching}: We follow the implementation details of flow matching described in \cite{black2024pi0visionlanguageactionflowmodel}. Compared with the standard flow matching approach, \citep{black2024pi0visionlanguageactionflowmodel} replaces the uniform sampling distribution of timesteps with a Beta distribution, which emphasizes lower (noisier) timesteps sampling. We adopt the same design and likewise use 10 integration steps in our implementation. \textbf{L1 Regression}: We adapt the L1 regression objectives to directly supervise the ground truth actions without other designs. 

\begin{table}[ht]
    \caption{\textbf{Overall results of the comparison with the standard methods.} Results show our method largely retains the advantages of flow matching, \(\ie\) fast convergence and better performance, with much fewer neural function evaluations (NFE), while mitigating the potential performance degradation associated with direct L1 regression.}
    \label{exp:mimicgen_all}
    \begin{center}
    \begin{small}
    \begin{sc}
    \renewcommand{\tabcolsep}{2pt}
    \begin{tabular}{c|ccc|c}
        \toprule
          Task&DP &Flow &L1&Ours\\
         \midrule
         Stack D1&\textbf{0.80}&\textbf{0.80}&0.74&\underline{0.78}\\
         Stack Three D1&\textbf{0.22}&\underline{0.20}&0.14&\underline{0.20}\\
         Threading D2&\underline{0.18}&0.14&0.12&\textbf{0.20}\\
         Square D2&0.06&\textbf{0.10}&0.06&\underline{0.08}\\
         Coffee D2&0.48&\textbf{0.52}&0.48&\underline{0.50}\\
         Hammer Cleanup D1 &\textbf{0.48}&\textbf{0.48}&\underline{0.46}&\textbf{0.48}\\
         Mug Cleanup D1&0.26&\textbf{0.30}&0.24&\underline{0.28}\\
         Nut Assembly D0&0.43&\underline{0.41}&0.35&0.40\\
         \midrule
         NFE&100&10&1&2\\
        \midrule
        Average.&0.364&\textbf{0.369}&0.324&\underline{0.365}\\
        \bottomrule
    \end{tabular}
    \end{sc}
    \end{small}
    \end{center}
\end{table}

Following the setting in \cite{chi2023diffusion}, all baselines share the same non-pretrained ResNet-18\cite{he2016deep} visual encoder and 1-D Conditional Temporal UNet~\cite{janner2022planning} architecture (except for L1 objective, which doesn't take the timestep as condition), while employing different objectives. We build the evaluation on 8 manipulation tasks in MimicGen~\cite{mandlekar2023mimicgen}.  We train all methods with the same 100 expert trajectories in relative control mode and report the best success rate until the current evaluation. The success rate of each evaluation is averaged over 50 different runs with varied random seeds. Figure \ref{fig:mimicgen} shows the trend of the best success rate over steps throughout the training process and Table \ref{exp:mimicgen_all} reports the overall results. The results indicate that flow matching exhibits better convergence efficiency and overall performance than diffusion across most tasks. Although L1 regression performs comparably to denoising-based methods in some cases and offers higher inference efficiency, it often suffers from significant performance degradation. L1 Flow achieves superior convergence efficiency and maintains performance comparable to or even surpassing of the majority of tasks with much less neural function evaluations (NFE), while also demonstrating a consistent performance advantage over L1 regression. The performance comparison with DDPM (100 steps) shows that our method largely retains the advantages of flow matching, while mitigating the potential performance degradation associated with direct L1 regression.

\subsection{Comparison with Other Fast Denoising Models}
\label{sec:comparison}

To further evaluate the performance and efficiency of our approach, we compare it with two representative fast denoising baselines, \textbf{Consistency Policy (CP)} \cite{prasad2024consistency} and \textbf{OneDP} \cite{wang2024one}. Both CP and OneDP are distillation-based methods that rely on a pretrained diffusion model. Specifically, CP depends on the boundary conditions of the EDM\cite{karras2022elucidating} noise scheduling, whereas OneDP is compatible with both DDPM and EDM schedulers. Therefore, diffusion policies with different schedulers (DDPM, DDIM, and EDM) are also included as baselines for comparison. In addition, OneDP includes two variants: a stochastic version (\textbf{OneDP-S}) and a deterministic version (\textbf{OneDP-D}). OneDP-S employs an auxiliary network to estimate the score of the generator distribution, enhancing the performance of the one-step policy in complex environments at the cost of a more computationally expensive training process. Conversely, OneDP-D removes the generator score network by directly optimizing a simplified score loss, yielding a deterministic observation-to-action policy. These two variants are included in the evaluation.

\begin{figure*}[t]
\centering
\begin{subfigure}{0.30\linewidth}
    \includegraphics[width=\linewidth]{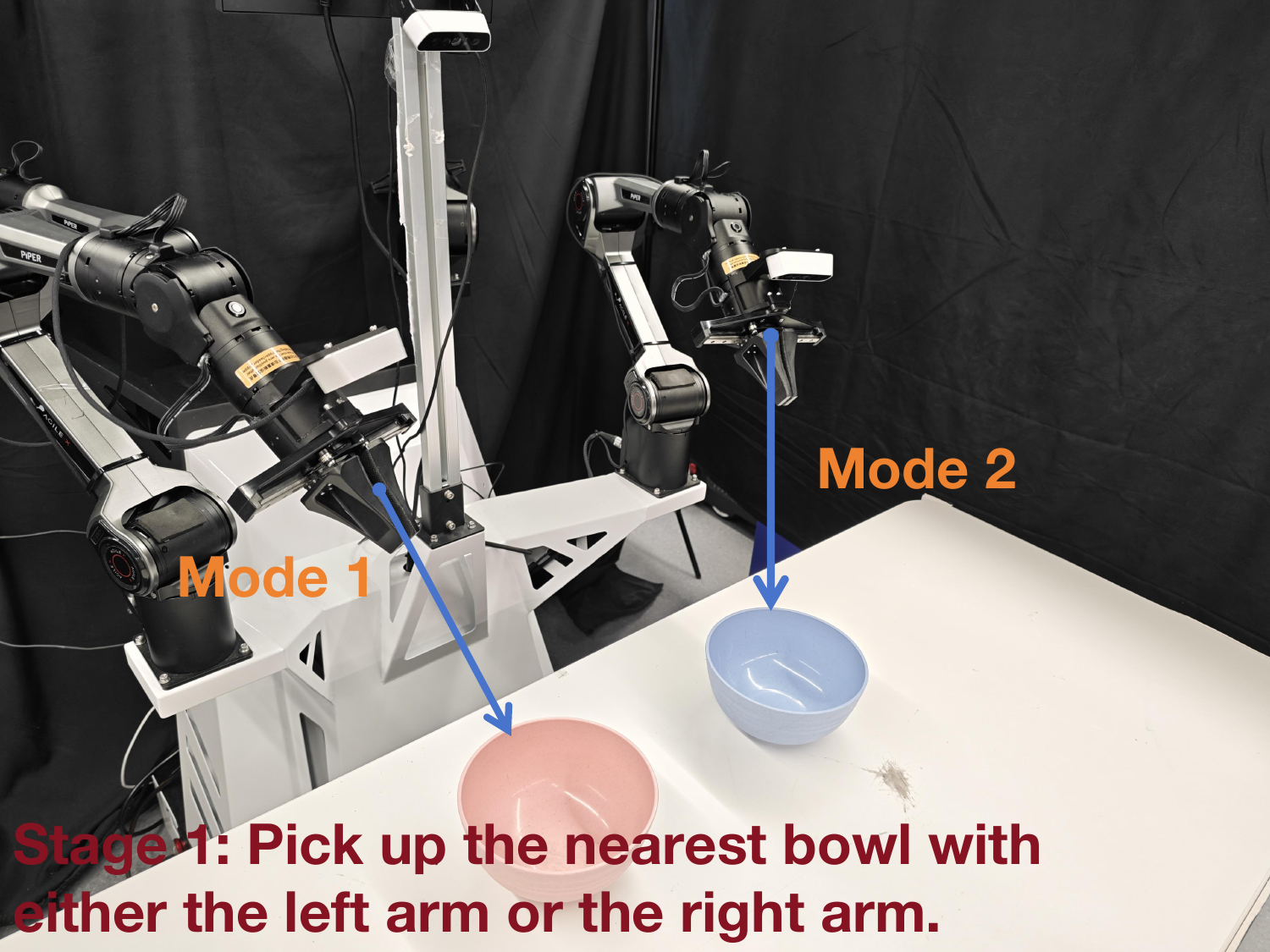}
    \caption{Stage 1}
    \label{fig:stage1}
  \end{subfigure}
\begin{subfigure}{0.30\linewidth}
    \includegraphics[width=\linewidth]{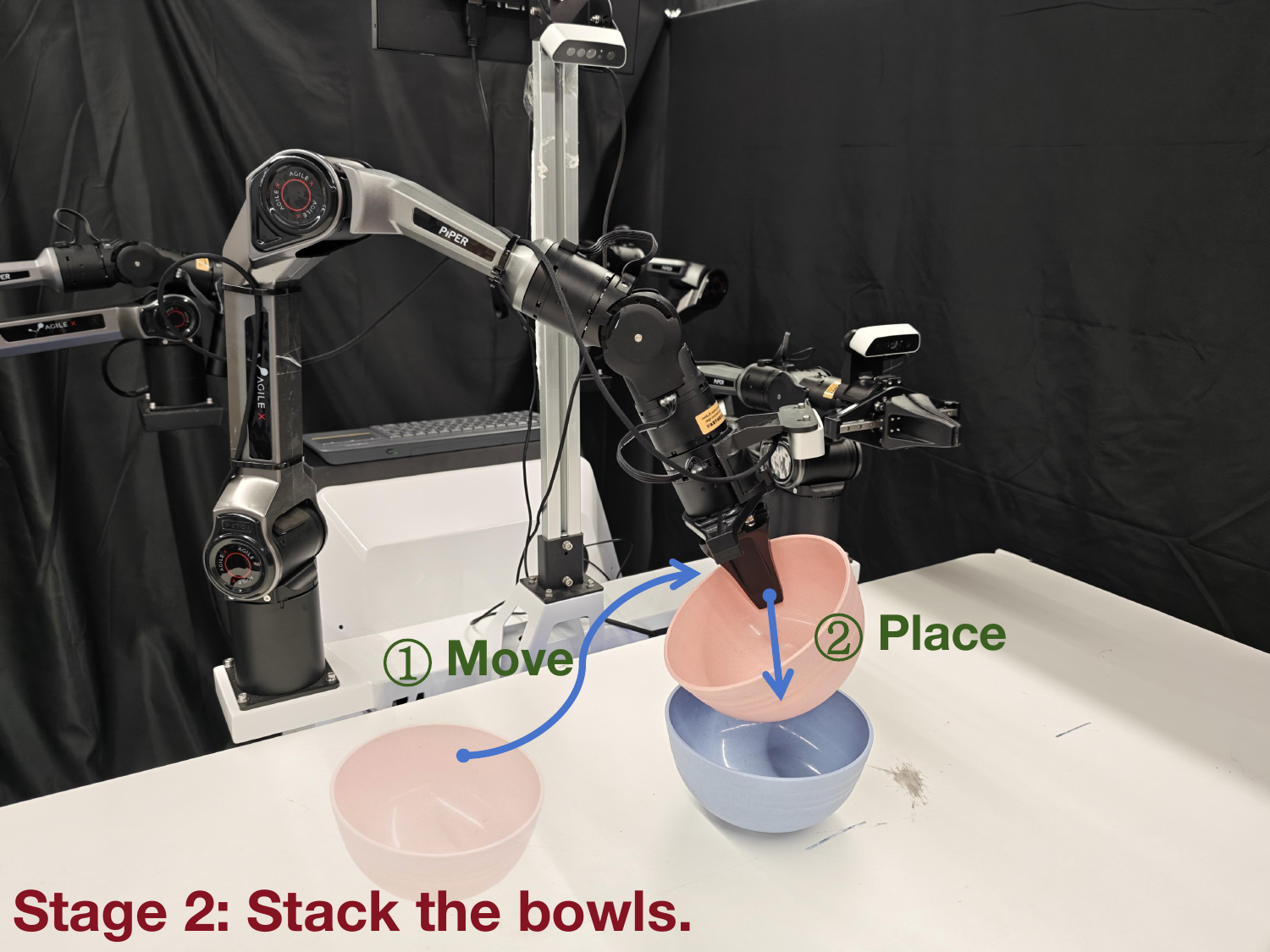}
    \caption{Stage 2}
    \label{fig:stage2}
  \end{subfigure}
  \begin{subfigure}{0.39\linewidth}
    \includegraphics[width=\linewidth]{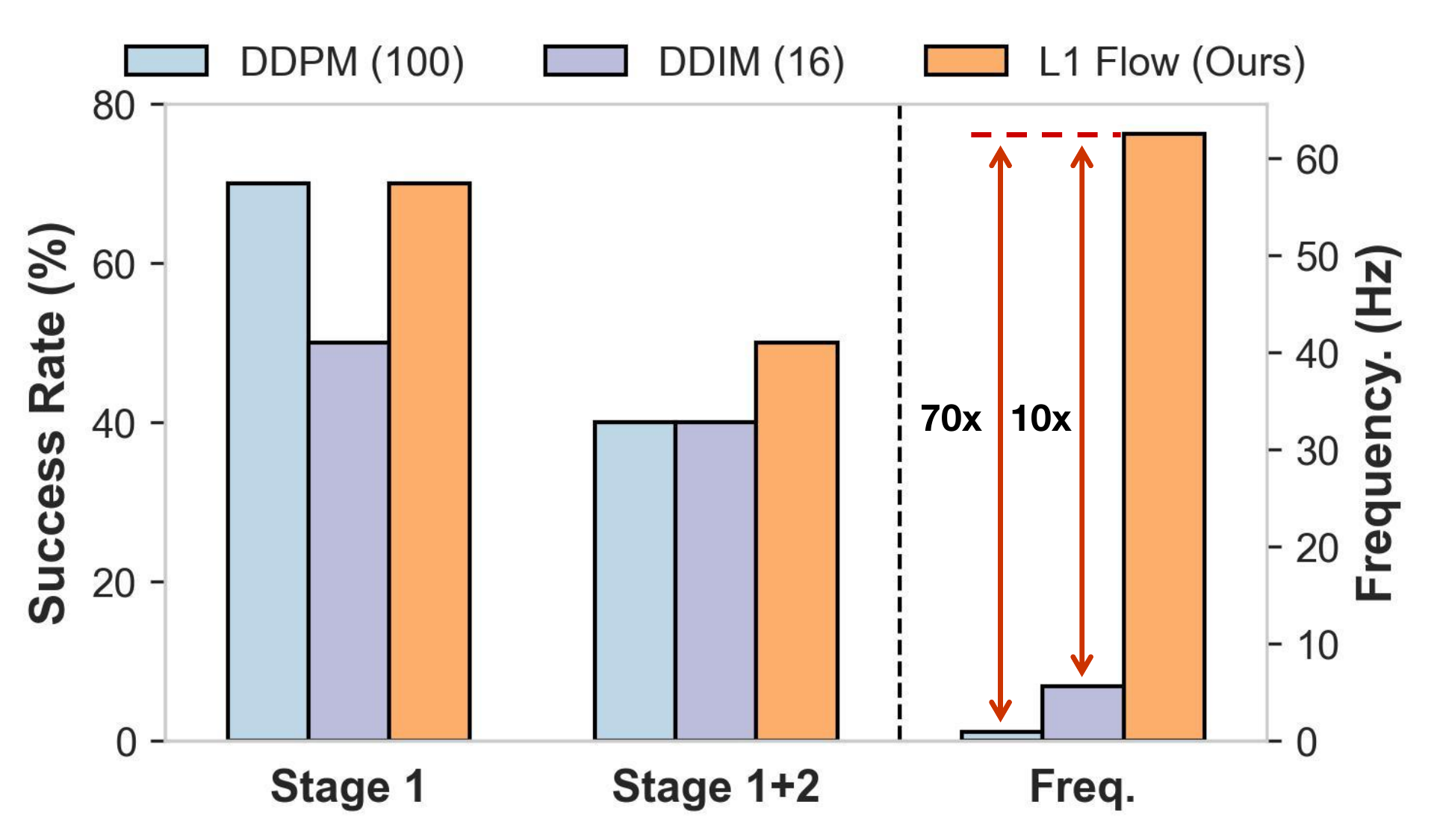}
    \caption{Overall Results}
    \label{fig:real exp}
  \end{subfigure}

\caption{\textbf{The real-world evaluation setup.} The evaluation is conducted in AGILEX Mobile Aloha platform, and we establish a task with two stages, emphasizing modeling the action multi-modality and prediction precision. 
\textbf{(a) Stage 1:} The robot is required to pick up the bowl with either the left or the right arm, requiring the policy to model the multi-modal action distribution. 
\textbf{(b) Stage 2:} The dual-arm is required to move the grasped bowl on top of the other bowls and place it, emphasizing the transformation from multi-modal to single-modal action prediction and also demands action precision, requiring the policy to accurately localize the bowl’s edge. 
\textbf{(c) Performance comparison with the standard diffusion policy}, including DDPM (100 steps) and DDIM (16 steps) in real-world task.
The results show that L1 Flow achieves comparable performance against the baselines and has a superior inference speed, achieving about \textbf{10-70\(\times\) speed-up}.} 
\label{fig:real-world}
\end{figure*}

Following the setting in OneDP, we conduct the evaluations on two established benchmarks, RoboMimic \cite{mandlekar2021matters} and PushT \cite{florence2022implicit}. We evaluate 5 challenging tasks in Robomimic with high-quality human demonstrations: Square-mh, Square-ph, ToolHang-ph, and Transport-ph, where ``ph'' denotes proficient human demonstrations and ``mh'' denotes mixed proficient/non-proficient human demonstrations. 
For PushT, we adopt the standard setting with 200 expert trajectories and use the RGB observations as input. During the evaluation, we report the peak success rate averaged over five different training runs and 100 random environment initializations (500 trials per task). The RoboMimic task metric is measured by binary success rates, while PushT is evaluated using goal-region coverage ranging from 0 to 1. The baseline results are taken from the original comparison in OneDP\cite{wang2024one} and marked with * in Table \ref{exp:robomimic}.

The baseline Diffusion Policy (DP) is trained for 1,000 epochs and sampled under DDPM, DDIM, and EDM noise scheduling. Both distillation-based methods require additional training stages: OneDP performs an extra 20 epochs for distillation, while CP requires around 450 additional epochs to converge. Accordingly, for distillation-based methods, the reported number of epochs includes both the pretraining epochs and the additional epochs required for the distillation. For comparison, our method is a single-stage approach trained from scratch with only 200 epochs.

In Table \ref{exp:robomimic}, our method achieves the highest success rates, outperforming the standard diffusion policies, CP, and OneDP in most tasks except ToolHang-ph. Although it shows degradation towards the initial DP in ToolHang-ph, it still outperforms CP and OneDP-D (EDM). In particular, it achieves strong performance with only two denoising steps (NFE=2) and only requires 200 epochs to converge, which is about 5\(\times\) to 7\(\times\) faster than the baselines.

\subsection{Real World Experiments}
In this section, we study the multi-modal action in real-world scenarios to evaluate the proposed method. Beyond the confirmed improvement (only 2 NFE) in inference latency, which is crucial for real-world deployment, we are also interested in whether this training objective offers other advantages or drawbacks in other aspects of the real deployment. Therefore, the task design primarily focuses on two topics: multi-modality modeling and precision single modality prediction.

\textbf{Platform.} 
Our experiments are conducted on the AGILEX ALOHA dual-arm platform, which includes two 6-DoF PIPER arms with grippers, two wrists, and one front-facing Orbbec DABAI depth camera.
All policies are executed on a PC workstation equipped with an NVIDIA RTX 4090 GPU and an Intel I7-13700 CPU.

\textbf{Task Definition.} The evaluation includes one task with 2 stages as follows. \textbf{Stage 1: Pick up the Bowl.} This task requires the dual-arm robot to use either left or right arm to pick up the nearest bowl on the table, \(\ie\) pick up the left bowl with the left arm or pick up the right bowls with arm arm. The provided demonstrations include an equal number of trajectories performed with the left and right arms, highlighting the policy’s capability to model multi-modal action distributions under the same observation. \textbf{Stage 2: Stack the Bowl.} The dual-arm robot is required to place the grasped bowl on the other bowl on the table. This stage highlights the common single modality modeling and the prediction precision, so as to avoid the accumulation of errors during execution. The overview of the task scene and setup is shown in Figure \ref{fig:stage1} and \ref{fig:stage2}.

\begin{table}[ht]
    \caption{\textbf{Ablation study on key components of the proposed method.} Results show the contributions of the training objectives and timestep sampling strategy to the overall performance. We also compare the performance across the initial flow matching with different integration steps. The best results are marked in \textbf{bold} and the second best is \underline{underlined}.}
    \label{exp:ablation}
    \centering
    \begin{scriptsize}
    \begin{sc}
    \renewcommand{\tabcolsep}{6pt}
    \begin{tabular}{l|c|cc}
        \toprule
        Method & NFE & PushT & Square-ph \\
        \midrule
        MSE& 2 & 0.821$\pm$0.012 \textcolor{red}{(-0.048)}& 0.916$\pm$0.009 \textcolor{red}{(-0.032)} \\
        \midrule
        Uniform & 2 & 0.854$\pm$0.033 \textcolor{red}{(-0.015)}& 0.928$\pm$0.011 \textcolor{red}{(-0.020)} \\
        Beta & 2 &  \underline{0.857$\pm$0.013} \textcolor{red}{(-0.012)}& 0.940$\pm$0.012 \textcolor{red}{(-0.008)} \\
        \midrule
        Flow & 1 & 0.755$\pm$0.023 \textcolor{red}{(-0.114)} & 0.928$\pm$0.013 \textcolor{red}{(-0.020)} \\
        Flow & 2 & 0.777$\pm$0.008 \textcolor{red}{(-0.092)} & \textbf{0.964$\pm$0.005} \textcolor{green}{(+0.016)} \\
        Flow & 10 & 0.812$\pm$0.015 \textcolor{red}{(-0.057)} & 0.946$\pm$0.011 \textcolor{red}{(-0.002)} \\
        Flow & 100 & 0.815$\pm$0.010 \textcolor{red}{(-0.054)} &  0.932$\pm$0.004 \textcolor{red}{(-0.016)} \\
        \midrule
        Ours & 2 & \textbf{0.869$\pm$0.007} & \underline{0.948$\pm$0.013} \\
        \bottomrule
    \end{tabular}
    \end{sc}
    \end{scriptsize}
\end{table}

\textbf{Details.} 
We collect 50 episodes via teleoperation for the task and record the data at 25 Hz. We keep most of the policy configurations used in the simulation and specifically make several modifications as follows: 
(1) input RGB images are resized to 224\(\times\)224 and without random crop, 
(2) action/proprioceptive state is defined in the 6D absolute joint-angle and 1D gripper space, 
(3) prediction action horizon is 32, and the execution action horizon is 16, 
(4) vision encoder is replaced by the pretrained ResNet18\cite{he2016deep} for better generalization in the real world. We train the diffusion policy for 200 epochs, while the L1 flow is trained for only 50 epochs for its fast convergence.

\textbf{Results.} 
Each result is obtained by averaging the
results of 10 executions. Figure~\ref{fig:real exp} shows that L1 Flow achieves comparable performance against the baselines, showcasing the potential of modeling multi-modal action distribution via this paradigm. 
We measure the computing time required for each method to generate one action chunk and convert it to the chunk frequency as shown in Figure \ref{fig:real exp}. Taking advantage of its 2-step denoising strategy and the simplification in each flow step compared with diffusion, L1 Flow achieves extremely fast inference speed, approximately 70\(\times\) faster than the original 100-step DDPM, and about 10\(\times\) faster than the accelerated 16-step DDIM.

\subsection{Ablation and Discussion}
\label{sec:ablation}

To further analyze the contribution of each component in our proposed framework, we conduct a comprehensive ablation study on two representative tasks, \textbf{PushT} and \textbf{Square-ph}, in Section \ref{sec:comparison}. All experiments are trained and evaluated in a consistent setting. Two key components are analyzed: \textbf{(1)} The impact of different regression objectives (L1 \(\vs\) MSE) and \textbf{(2)} The effect of different timestep sampling strategies (Uniform \(\vs\) Beta \(\vs\) Logistic Normal Mixture). In addition, we also include the discussion of \textbf{(3)} the comparison between the standard flow matching with different integration steps and our proposed 2-step L1 Flow. The results are summarized in Table~\ref{exp:ablation}. 

\subsubsection{L1 vs MSE Loss}
We first compare the L1 and MSE training objective under identical configurations. As the L1 training objective is increasingly adopted by large vision-language-action models\cite{kim2025fine,wang2025vla} for direct action regression, we explore replacing the commonly used MSE loss for noise or velocity regression in denoising-based methods with L1 loss to supervise the ground-truth samples in our sample-prediction flow matching. As shown in Table~\ref{exp:ablation}, the L1 objective yields consistently higher success rates on both tasks. The empirical results shown in \cite{kim2025fine,wang2025vla}, along with our comparison, suggest the advantages of the L1 objective in supervising the ground-truth sample. However, it is still unclear where this superiority stems from. One of the possible future directions will focus on delving into the distributional difference between sample prediction and noise-like prediction, which could offer guidance for choosing more appropriate learning objectives rather than empirically.

\subsubsection{Uniform vs Beta vs Mixed Logistic Normal}
To investigate the effect of the timestep sampling strategy, we compare the original \textit{Uniform distribution} in flow matching and \textit{Beta distribution} used in \cite{black2024pi0visionlanguageactionflowmodel} with our applied \textit{Mixed Logistic Normal Distribution} (Equation \ref{eq:mlog}). Specifically, the \textit{Uniform distribution} samples timesteps evenly across the entire range, the \textit{Beta strategy} places greater emphasis on high-noise level, and our \textit{Mixed Logistic Normal} prioritizes sampling around the intermediate timesteps. The results show that replacing the \textit{Mixed Logistic Normal} with the other two leads to varying degrees of performance degradation on both tasks. The variation indicates that emphasizing intermediate timesteps can benefit our two-step sampling.

\subsubsection{L1 Flow vs Standard Flow Matching}
We compare our method with standard flow matching under varying inference budgets. In \textbf{PushT}, 1-step flow matching exhibits non-trivial performance compared with 1 step diffusion shown in Table \ref{exp:robomimic}, and there are performance gains with increasing number of integration steps. In \textbf{Square-ph}, standard flow matching gains the best success rate in a 2-step setting and shows degradation as the integration step number increases, which is likely due to the accumulation error. This indicates that the optimal inference budget for flow matching varies widely between different task settings. In comparison, our method exhibits highly competitive performance in both tasks with fixed 2-step sampling. Therefore, from another perspective, our method can be viewed as a stable alternative that avoids the need for dynamically selecting the number of integration steps based on the task while simultaneously addressing the performance degradation caused by a small number of steps and the accumulated error arising from a large number of steps.

\subsubsection{Timestep Strategy}
\label{sec:inference_ablation}

We examines the empirical optimality of targeting the temporal midpoint in the first flow integration. In our two-step denoising schedule, inference consists of two stages: the first is to perform one step integration to a selective time point $t_{\mathrm{first}}$, followed by direct model prediction at $t_{\mathrm{first}}$ to obtain the final sample $x_1$. We sweep $t_{\mathrm{first}} \in [0.1, 0.9]$ to empirically verify whether the temporal midpoint ($t_{\mathrm{first}} = 0.5$) indeed yields optimal performance. The second study analyzes the efficiency of our 2-step sample approach relative to the standard integration inference paradigm with our sample prediction flow matching, which integrates with different step sizes \(\ie\) NFE $\in \{1, 2, 5, 10\}$. All models are trained using the default L1 Flow configuration on the PushT task.

Results in Table~\ref{tab:inference_ablation} reveal two insights: 

(1) Our two-step inference achieves best performance at $t_{\mathrm{first}} = 0.5$, confirming the empirical optimality of the temporal midpoint in our method. Notably, it outperforms all standard multi-step baselines, demonstrating that our two-step design eliminates redundant integration steps while preserving accuracy. This validates that our strategy efficiently captures optimal flow behavior with minimal computational overhead. 

(2) Among standard baselines, NFE=2 yields the highest success rate ($0.863 \pm 0.008$), closely approaching our method’s best ($0.869 \pm 0.007$). This is expected, the standard NFE=2 integration is mathematically equivalent to our two-step schedule under exact ODE solving. The minor performance gap ($\sim$0.006) therefore arises primarily from numerical integration errors in the calculation of the velocity, whereas our approach bypasses intermediate integration and directly predicts $x_1$.

\begin{table}[!htbp]
\centering
\caption{\textbf{Ablation study on timestep strategy.} All models are trained using the default L1 Flow configuration on the \textbf{PushT} task. ``Ours'' means using our proposed two-step denoising schedule, while ``Standard'' refers to the conventional multi-step Flow Matching inference method. We report the mean and standard deviation of success rates over five independent training runs, each evaluated under 100 randomized environment initializations (500 trials in total). The best results are marked in \textbf{bold} and the second best is \underline{underlined}.}
\label{tab:inference_ablation}
\begin{tabular}{lccc}
\toprule
\textbf{Strategy} & $\mathbf{t}_{\mathrm{first}}$ & \textbf{NFE} & \textbf{PushT} \\
\midrule
\multirow{9}{*}{Ours} 
& 0.1 & \multirow{9}{*}{2} & 0.846 $\pm$ 0.022 \\
& 0.2 & & 0.853 $\pm$ 0.016 \\
& 0.3 & & 0.843 $\pm$ 0.009 \\
& 0.4 & & 0.843 $\pm$ 0.020 \\
& 0.5 & & \textbf{0.869 $\pm$ 0.007} \\
& 0.6 & & 0.855 $\pm$ 0.012 \\
& 0.7 & & 0.855 $\pm$ 0.018 \\
& 0.8 & & 0.849 $\pm$ 0.019 \\
& 0.9 & & 0.848 $\pm$ 0.017 \\
\cmidrule(lr){1-4}
\multirow{4}{*}{Standard} 
& \multicolumn{1}{c}{---} & 1 & 0.845 $\pm$ 0.017 \\
& \multicolumn{1}{c}{---} & 2 & \underline{0.863 $\pm$ 0.008} \\
& \multicolumn{1}{c}{---} & 5 & 0.847 $\pm$ 0.016 \\
& \multicolumn{1}{c}{---} & 10 & 0.856 $\pm$ 0.007 \\
\bottomrule
\end{tabular}
\end{table}

\begin{table}[!t]
\centering
\caption{\textbf{Ablation study on loss type.} (v-loss vs. x-loss) under the same configuration as Table~\ref{tab:inference_ablation}, where ``2-step'' adopts $t_{\mathrm{first}} = 0.5$. ``v-loss'' and ``x-loss'' denote losses computed in velocity space and action($x$) space, respectively, while both targeting action prediction. For each loss type, the best results are marked in \textbf{bold} and the second best is \underline{underlined}.}
\label{tab:loss_ablation}
\begin{tabular}{cccc}
\toprule
\textbf{Loss Type} & \textbf{Strategy} & \textbf{NFE} & \textbf{PushT} \\
\midrule
\multirow{5}{*}{x-loss (MSE)} 
& Standard       & 1  & \underline{0.842 $\pm$ 0.014} \\
& Standard       & 2  & 0.817 $\pm$ 0.014 \\
& Standard       & 5  & 0.836 $\pm$ 0.013 \\
& Standard       & 10 & \textbf{0.843 $\pm$ 0.013} \\
& 2-step   & 2  & 0.821 $\pm$ 0.012 \\
\cmidrule(lr){1-4}
\multirow{5}{*}{v-loss (MSE)} 
& Standard       & 1  & \textbf{0.851 $\pm$ 0.023} \\
& Standard       & 2  & 0.832 $\pm$ 0.031 \\
& Standard       & 5  & 0.844 $\pm$ 0.010 \\
& Standard       & 10 & \underline{0.850 $\pm$ 0.020} \\
& 2-step   & 2  & 0.847 $\pm$ 0.017 \\
\cmidrule(lr){1-4}
\multirow{5}{*}{x-loss (L1)} 
& Standard       & 1  & 0.845 $\pm$ 0.017 \\
& Standard       & 2  & \underline{0.863 $\pm$ 0.008} \\
& Standard       & 5  & 0.847 $\pm$ 0.016 \\
& Standard       & 10 & 0.856 $\pm$ 0.007 \\
& 2-step (Ours)  & 2  & \textbf{0.869 $\pm$ 0.007} \\
\cmidrule(lr){1-4}
\multirow{5}{*}{v-loss (L1)} 
& Standard       & 1  & 0.865 $\pm$ 0.012 \\
& Standard       & 2  & 0.865 $\pm$ 0.019 \\
& Standard       & 5  & \underline{0.870 $\pm$ 0.014} \\
& Standard       & 10 & \textbf{0.872 $\pm$ 0.018} \\
& 2-step  & 2  &0.868 $\pm$ 0.013 \\
\bottomrule
\end{tabular}
\end{table}

\subsubsection{Loss Formulation}
\label{sec:loss_ablation}

Recently, \citep{li2025jit} has also applied the sample-prediction variant of flow matching in the image generation domain, while adopting to regress velocity as the final objective. To examine how different loss formulations affect learning dynamics and downstream control performance, we perform ablation studies comparing velocity-space loss ($v$-loss) and sample-space loss ($x$-loss) under identical experimental settings. 

The sample-preditcion model directly predicts the target sample $x_1$ and the sample-space loss ($x$-loss) is naturally obtained as

\begin{equation}
\mathcal{L}_x = \mathbb{E}_{t, x_0, x_1} \left\| f_\theta(x_t, t) - x_1 \right\|
\end{equation}

Following the notation in Section~\ref{sec:methodology}, we denote $x_0$ as Gaussian noise, $x_1$ as the ground-truth target sample, and define the linear interpolation path $x_t = t x_1 + (1-t)x_0$. Along this trajectory, the ground-truth velocity field is as follows:

\begin{equation}
v_{\text{gt}} = x_1 - x_0 = \frac{x_1 - x_t}{1 - t}.
\label{eq:v_truth}
\end{equation}

Because the model predicts $x_1$ directly (denoted as $f_\theta(x_t,t)$), the induced velocity prediction becomes

\begin{equation}
v_{\text{pred}} = \frac{f_\theta(x_t, t) - x_t}{1 - t}.
\label{eq:v_pred}
\end{equation}

The velocity-space loss is thus given by the discrepancy between predicted and ground-truth velocities:

\begin{align}
    \mathcal{L}_v 
    &= \mathbb{E}_{t, x_0, x_1} \left\| v_{\text{pred}} - v_{\text{gt}} \right\|  \nonumber \\
    &= \mathbb{E}_{t, x_0, x_1}  \left\| \frac{f_\theta(x_t, t) - x_1}{1-t} \right\|.
    \label{eq:v_loss_L1}
\end{align}

This derivation reveals that $\mathcal{L}_v$ is essentially a time-dependent reweighting of $\mathcal{L}_x$, where the factor $\frac{1}{1-t}$ increasingly amplifies prediction errors as $t \rightarrow 1$, \(\ie\), the low-level noisy sample.

All experiments follow the configuration described in Section~\ref{sec:inference_ablation}. We evaluate four loss variants, combining velocity-space loss ($v$-loss) and sample-space loss ($x$-loss) with either L1 Loss or Mean Squared Error (MSE). Furthermore, we conduct comprehensive evaluations across multiple inference strategies mentioned in Section~\ref{sec:inference_ablation}, and the full results are summarized in Table~\ref{tab:loss_ablation}.

Experiments on PushT yield three main observations:

(1) \textbf{L1 loss consistently outperform their MSE counterparts}. It is still unclear where this superiority stems from, but we recommend L1 loss as the default loss for sample prediction flow matching. MSE remains a viable alternative when smooth gradient signals are prioritized, though it generally yields slightly lower robustness in our experiments.

(2) \textbf{The $v$-loss generally achieves higher best performance than $x$-loss.} Specifically, $v$-loss (L1) attains the highest overall success rate (0.872 at NFE=10), surpassing our method, $x$-loss (L1) with 2-step inference (0.869), by a marginal +0.003. Nevertheless, our approach offers improved inference efficiency (5× fewer function evaluations), illustrating a clear trade-off between maximum accuracy and practical deployment speed.

(3) \textbf{The one-step prediction of sample-prediction flow matching yields a non-trivial outcome.} Under the same setting, the one-step prediction performance of sample-prediction flow matching is substantially higher than that of v-prediction (refer to Table \ref{exp:ablation}) and in some cases even surpasses the multi-step prediction results. This suggests that multi-step denoising may be redundant for robotic action modeling, and that understanding how to booster one-step prediction to its optimal performance could be an intriguing direction for future research.

\section{Conclusion}
This work targets at the key trade-off between multi-modal distribution modeling capability and efficiency in visuomotor policy learning. We propose L1 Flow, a novel framework that reforms velocity-prediction flow matching into a sample-prediction paradigm with an L1 training objective, enabling the fusion of denoising models’ strengths and L1 regression’s efficiency. It achieves comparable or better performance than standard denoising-based methods (e.g., diffusion, flow matching) with 10–70× accelerated inference and distillation-based approaches (e.g., Consistency Policy, OneDP) with 5–7× faster training convergence, which provides a practical alternative for real-time robotic manipulation by balancing expressiveness and efficiency.

{
    \small
    \bibliographystyle{ieeenat_fullname}
    \bibliography{main}

@String(AAAI = {AAAI})

@article{li2025jit,
  title={Back to Basics: Let Denoising Generative Models Denoise},
  author={Li, Tianhong and He, Kaiming},
  journal={arXiv preprint arXiv:2511.13720},
  year={2025}
}

@article{su2025dense,
  title={Dense policy: Bidirectional autoregressive learning of actions},
  author={Su, Yue and Zhan, Xinyu and Fang, Hongjie and Xue, Han and Fang, Hao-Shu and Li, Yong-Lu and Lu, Cewu and Yang, Lixin},
  journal={arXiv preprint arXiv:2503.13217},
  year={2025}
}

@inproceedings{gong2025carp,
  title={Carp: Visuomotor policy learning via coarse-to-fine autoregressive prediction},
  author={Gong, Zhefei and Ding, Pengxiang and Lyu, Shangke and Huang, Siteng and Sun, Mingyang and Zhao, Wei and Fan, Zhaoxin and Wang, Donglin},
  booktitle={Proceedings of the IEEE/CVF International Conference on Computer Vision},
  pages={13460--13470},
  year={2025}
}

@inproceedings{
sun2025scorebased,
title={Score-Based Diffusion Policy Compatible with Reinforcement Learning via Optimal Transport},
author={Mingyang Sun and Pengxiang Ding and Weinan Zhang and Donglin Wang},
booktitle={Forty-second International Conference on Machine Learning},
year={2025},
url={https://openreview.net/forum?id=2dqiqST8ZJ}
}

@inproceedings{
wang2023diffusion,
title={Diffusion Policies as an Expressive Policy Class for Offline Reinforcement Learning},
author={Zhendong Wang and Jonathan J Hunt and Mingyuan Zhou},
booktitle={The Eleventh International Conference on Learning Representations },
year={2023},
url={https://openreview.net/forum?id=AHvFDPi-FA}
}

@inproceedings{
park2025flow,
title={Flow Q-Learning},
author={Seohong Park and Qiyang Li and Sergey Levine},
booktitle={Forty-second International Conference on Machine Learning},
year={2025},
url={https://openreview.net/forum?id=KVf2SFL1pi}
}

@inproceedings{
ren2025diffusion,
title={Diffusion Policy Policy Optimization},
author={Allen Z. Ren and Justin Lidard and Lars Lien Ankile and Anthony Simeonov and Pulkit Agrawal and Anirudha Majumdar and Benjamin Burchfiel and Hongkai Dai and Max Simchowitz},
booktitle={The Thirteenth International Conference on Learning Representations},
year={2025},
url={https://openreview.net/forum?id=mEpqHvbD2h}
}

@article{karras2022elucidating,
  title={Elucidating the design space of diffusion-based generative models},
  author={Karras, Tero and Aittala, Miika and Aila, Timo and Laine, Samuli},
  journal={Advances in neural information processing systems},
  volume={35},
  pages={26565--26577},
  year={2022}
}

@inproceedings{Song2023ConsistencyM,
  title={Consistency Models},
  author={Yang Song and Prafulla Dhariwal and Mark Chen and Ilya Sutskever},
  booktitle={International Conference on Machine Learning},
  year={2023},
  url={https://api.semanticscholar.org/CorpusID:257280191}
}

@inproceedings{
song2024improved,
title={Improved Techniques for Training Consistency Models},
author={Yang Song and Prafulla Dhariwal},
booktitle={The Twelfth International Conference on Learning Representations},
year={2024},
url={https://openreview.net/forum?id=WNzy9bRDvG}
}

@inproceedings{
kim2024consistency,
title={Consistency Trajectory Models: Learning Probability Flow {ODE} Trajectory of Diffusion},
author={Dongjun Kim and Chieh-Hsin Lai and Wei-Hsiang Liao and Naoki Murata and Yuhta Takida and Toshimitsu Uesaka and Yutong He and Yuki Mitsufuji and Stefano Ermon},
booktitle={The Twelfth International Conference on Learning Representations},
year={2024},
url={https://openreview.net/forum?id=ymjI8feDTD}
}

@article{prasad2024consistency,
  title={Consistency policy: Accelerated visuomotor policies via consistency distillation},
  author={Prasad, Aaditya and Lin, Kevin and Wu, Jimmy and Zhou, Linqi and Bohg, Jeannette},
  journal={arXiv preprint arXiv:2405.07503},
  year={2024}
}

@inproceedings{he2016deep,
  title={Deep residual learning for image recognition},
  author={He, Kaiming and Zhang, Xiangyu and Ren, Shaoqing and Sun, Jian},
  booktitle={Proceedings of the IEEE conference on computer vision and pattern recognition},
  pages={770--778},
  year={2016}
}

@article{janner2022planning,
  title={Planning with diffusion for flexible behavior synthesis},
  author={Janner, Michael and Du, Yilun and Tenenbaum, Joshua B and Levine, Sergey},
  journal={arXiv preprint arXiv:2205.09991},
  year={2022}
}

@article{mandlekar2023mimicgen,
  title={Mimicgen: A data generation system for scalable robot learning using human demonstrations},
  author={Mandlekar, Ajay and Nasiriany, Soroush and Wen, Bowen and Akinola, Iretiayo and Narang, Yashraj and Fan, Linxi and Zhu, Yuke and Fox, Dieter},
  journal={arXiv preprint arXiv:2310.17596},
  year={2023}
}

@article{mandlekar2021matters,
  title={What matters in learning from offline human demonstrations for robot manipulation},
  author={Mandlekar, Ajay and Xu, Danfei and Wong, Josiah and Nasiriany, Soroush and Wang, Chen and Kulkarni, Rohun and Fei-Fei, Li and Savarese, Silvio and Zhu, Yuke and Mart{\'\i}n-Mart{\'\i}n, Roberto},
  journal={arXiv preprint arXiv:2108.03298},
  year={2021}
}

@article{zhao2023learning,
  title={Learning fine-grained bimanual manipulation with low-cost hardware},
  author={Zhao, Tony Z and Kumar, Vikash and Levine, Sergey and Finn, Chelsea},
  journal={arXiv preprint arXiv:2304.13705},
  year={2023}
}

@article{geng2025mean,
  title={Mean flows for one-step generative modeling},
  author={Geng, Zhengyang and Deng, Mingyang and Bai, Xingjian and Kolter, J Zico and He, Kaiming},
  journal={arXiv preprint arXiv:2505.13447},
  year={2025}
}

@article{zou2025dm1,
  title={DM1: MeanFlow with Dispersive Regularization for 1-Step Robotic Manipulation},
  author={Zou, Guowei and Wang, Haitao and Wu, Hejun and Qian, Yukun and Wang, Yuhang and Li, Weibing},
  journal={arXiv preprint arXiv:2510.07865},
  year={2025}
}

@article{shukor2025smolvla,
  title={Smolvla: A vision-language-action model for affordable and efficient robotics},
  author={Shukor, Mustafa and Aubakirova, Dana and Capuano, Francesco and Kooijmans, Pepijn and Palma, Steven and Zouitine, Adil and Aractingi, Michel and Pascal, Caroline and Russi, Martino and Marafioti, Andres and others},
  journal={arXiv preprint arXiv:2506.01844},
  year={2025}
}

@misc{intelligence2025pi05visionlanguageactionmodelopenworld,
      title={$\pi_{0.5}$: a Vision-Language-Action Model with Open-World Generalization}, 
      author={Physical Intelligence and Kevin Black and Noah Brown and James Darpinian and Karan Dhabalia and Danny Driess and Adnan Esmail and Michael Equi and Chelsea Finn and Niccolo Fusai and Manuel Y. Galliker and Dibya Ghosh and Lachy Groom and Karol Hausman and Brian Ichter and Szymon Jakubczak and Tim Jones and Liyiming Ke and Devin LeBlanc and Sergey Levine and Adrian Li-Bell and Mohith Mothukuri and Suraj Nair and Karl Pertsch and Allen Z. Ren and Lucy Xiaoyang Shi and Laura Smith and Jost Tobias Springenberg and Kyle Stachowicz and James Tanner and Quan Vuong and Homer Walke and Anna Walling and Haohuan Wang and Lili Yu and Ury Zhilinsky},
      year={2025},
      eprint={2504.16054},
      archivePrefix={arXiv},
      primaryClass={cs.LG},
      url={https://arxiv.org/abs/2504.16054}, 
}

@article{liu2022flow,
  title={Flow straight and fast: Learning to generate and transfer data with rectified flow},
  author={Liu, Xingchao and Gong, Chengyue and Liu, Qiang},
  journal={arXiv preprint arXiv:2209.03003},
  year={2022}
}

@article{lipman2022flow,
  title={Flow matching for generative modeling},
  author={Lipman, Yaron and Chen, Ricky TQ and Ben-Hamu, Heli and Nickel, Maximilian and Le, Matt},
  journal={arXiv preprint arXiv:2210.02747},
  year={2022}
}

@article{wang2025vla,
  title={VLA-Adapter: An Effective Paradigm for Tiny-Scale Vision-Language-Action Model},
  author={Wang, Yihao and Ding, Pengxiang and Li, Lingxiao and Cui, Can and Ge, Zirui and Tong, Xinyang and Song, Wenxuan and Zhao, Han and Zhao, Wei and Hou, Pengxu and others},
  journal={arXiv preprint arXiv:2509.09372},
  year={2025}
}

@software{rdt2,
    title={RDT2: Enabling Zero-Shot Cross-Embodiment Generalization by Scaling Up UMI Data},
    author={RDT Team},
    url={https://github.com/thu-ml/RDT2},
    month={September},
    year={2025}
}

@misc{black2024pi0visionlanguageactionflowmodel,
      title={$\pi_0$: A Vision-Language-Action Flow Model for General Robot Control}, 
      author={Kevin Black and Noah Brown and Danny Driess and Adnan Esmail and Michael Equi and Chelsea Finn and Niccolo Fusai and Lachy Groom and Karol Hausman and Brian Ichter and Szymon Jakubczak and Tim Jones and Liyiming Ke and Sergey Levine and Adrian Li-Bell and Mohith Mothukuri and Suraj Nair and Karl Pertsch and Lucy Xiaoyang Shi and James Tanner and Quan Vuong and Anna Walling and Haohuan Wang and Ury Zhilinsky},
      year={2024},
      eprint={2410.24164},
      archivePrefix={arXiv},
      primaryClass={cs.LG},
      url={https://arxiv.org/abs/2410.24164}, 
}

@inproceedings{zhang2025flowpolicy,
  title={Flowpolicy: Enabling fast and robust 3d flow-based policy via consistency flow matching for robot manipulation},
  author={Zhang, Qinglun and Liu, Zhen and Fan, Haoqiang and Liu, Guanghui and Zeng, Bing and Liu, Shuaicheng},
  booktitle={Proceedings of the AAAI Conference on Artificial Intelligence},
  volume={39},
  number={14},
  pages={14754--14762},
  year={2025}
}

@article{wang2024one,
  title={One-step diffusion policy: Fast visuomotor policies via diffusion distillation},
  author={Wang, Zhendong and Li, Zhaoshuo and Mandlekar, Ajay and Xu, Zhenjia and Fan, Jiaojiao and Narang, Yashraj and Fan, Linxi and Zhu, Yuke and Balaji, Yogesh and Zhou, Mingyuan and others},
  journal={arXiv preprint arXiv:2410.21257},
  year={2024}
}

@article{kim2025fine,
  title={Fine-tuning vision-language-action models: Optimizing speed and success},
  author={Kim, Moo Jin and Finn, Chelsea and Liang, Percy},
  journal={arXiv preprint arXiv:2502.19645},
  year={2025}
}

@inproceedings{
wen2025diffusionvla,
title={Diffusion{VLA}: Scaling Robot Foundation Models via Unified Diffusion and Autoregression},
author={Junjie Wen and Yichen Zhu and Minjie Zhu and Zhibin Tang and Jinming Li and Zhongyi Zhou and Xiaoyu Liu and Chaomin Shen and Yaxin Peng and Feifei Feng},
booktitle={Forty-second International Conference on Machine Learning},
year={2025},
url={https://openreview.net/forum?id=VdwdU81Uzy}
}

@article{wen2025dexvla,
  title={Dexvla: Vision-language model with plug-in diffusion expert for general robot control},
  author={Wen, Junjie and Zhu, Yichen and Li, Jinming and Tang, Zhibin and Shen, Chaomin and Feng, Feifei},
  journal={arXiv preprint arXiv:2502.05855},
  year={2025}
}

@article{wen2025tinyvla,
  title={Tinyvla: Towards fast, data-efficient vision-language-action models for robotic manipulation},
  author={Wen, Junjie and Zhu, Yichen and Li, Jinming and Zhu, Minjie and Tang, Zhibin and Wu, Kun and Xu, Zhiyuan and Liu, Ning and Cheng, Ran and Shen, Chaomin and others},
  journal={IEEE Robotics and Automation Letters},
  year={2025},
  publisher={IEEE}
}

@article{wang2024sparse,
  title={Sparse diffusion policy: A sparse, reusable, and flexible policy for robot learning},
  author={Wang, Yixiao and Zhang, Yifei and Huo, Mingxiao and Tian, Ran and Zhang, Xiang and Xie, Yichen and Xu, Chenfeng and Ji, Pengliang and Zhan, Wei and Ding, Mingyu and others},
  journal={arXiv preprint arXiv:2407.01531},
  year={2024}
}

@article{ze20243d,
  title={3d diffusion policy: Generalizable visuomotor policy learning via simple 3d representations},
  author={Ze, Yanjie and Zhang, Gu and Zhang, Kangning and Hu, Chenyuan and Wang, Muhan and Xu, Huazhe},
  journal={arXiv preprint arXiv:2403.03954},
  year={2024}
}

@inproceedings{florence2022implicit,
  title={Implicit behavioral cloning},
  author={Florence, Pete and Lynch, Corey and Zeng, Andy and Ramirez, Oscar A and Wahid, Ayzaan and Downs, Laura and Wong, Adrian and Lee, Johnny and Mordatch, Igor and Tompson, Jonathan},
  booktitle={Conference on robot learning},
  pages={158--168},
  year={2022},
  organization={PMLR}
}

@article{shafiullah2022behavior,
  title={Behavior transformers: Cloning $ k $ modes with one stone},
  author={Shafiullah, Nur Muhammad and Cui, Zichen and Altanzaya, Ariuntuya Arty and Pinto, Lerrel},
  journal={Advances in neural information processing systems},
  volume={35},
  pages={22955--22968},
  year={2022}
}

@article{wang2024equivariant,
  title={Equivariant diffusion policy},
  author={Wang, Dian and Hart, Stephen and Surovik, David and Kelestemur, Tarik and Huang, Haojie and Zhao, Haibo and Yeatman, Mark and Wang, Jiuguang and Walters, Robin and Platt, Robert},
  journal={arXiv preprint arXiv:2407.01812},
  year={2024}
}

@article{liu2024rdt,
  title={Rdt-1b: a diffusion foundation model for bimanual manipulation},
  author={Liu, Songming and Wu, Lingxuan and Li, Bangguo and Tan, Hengkai and Chen, Huayu and Wang, Zhengyi and Xu, Ke and Su, Hang and Zhu, Jun},
  journal={arXiv preprint arXiv:2410.07864},
  year={2024}
}

@article{chi2023diffusion,
  title={Diffusion policy: Visuomotor policy learning via action diffusion},
  author={Chi, Cheng and Xu, Zhenjia and Feng, Siyuan and Cousineau, Eric and Du, Yilun and Burchfiel, Benjamin and Tedrake, Russ and Song, Shuran},
  journal={The International Journal of Robotics Research},
  pages={02783649241273668},
  year={2023},
  publisher={SAGE Publications Sage UK: London, England}
}

@article{song2020denoising,
  title={Denoising diffusion implicit models},
  author={Song, Jiaming and Meng, Chenlin and Ermon, Stefano},
  journal={arXiv preprint arXiv:2010.02502},
  year={2020}
}
}

\clearpage
\setcounter{page}{1}
\maketitlesupplementary
In the supplementary material, the additional experimental details with regard to 2 simulation experiments and the real-world experiment.

\section{Experiment Details}
\label{sec:imple_details}
\subsection{MimicGen}
\subsubsection{Simulation Setup}
\begin{figure*}[t]
  \centering
  \begin{subfigure}{0.2\linewidth}
    \includegraphics[width=\linewidth]{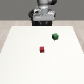}
    \caption{Stack D1}
  \end{subfigure}
  \hfill
  \begin{subfigure}{0.2\linewidth}
    \includegraphics[width=\linewidth]{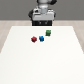}
    \caption{Stack Three D1}
  \end{subfigure}
  \hfill
  \begin{subfigure}{0.2\linewidth}
    \includegraphics[width=\linewidth]{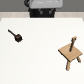}
    \caption{Threading D2}
  \end{subfigure}
  \hfill
  \begin{subfigure}{0.2\linewidth}
    \includegraphics[width=\linewidth]{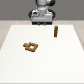}
    \caption{Square D2}
  \end{subfigure}
  
  \vspace{2pt}
  
  \begin{subfigure}{0.2\linewidth}
    \includegraphics[width=\linewidth]{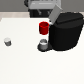}
    \caption{Coffee D2}
  \end{subfigure}
  \hfill
  \begin{subfigure}{0.2\linewidth}
    \includegraphics[width=\linewidth]{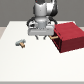}
    \caption{Hammer Cleanup D1}
  \end{subfigure}
  \hfill
  \begin{subfigure}{0.2\linewidth}
    \includegraphics[width=\linewidth]{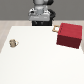}
    \caption{Mug Cleanup D1}
  \end{subfigure}
  \hfill
  \begin{subfigure}{0.2\linewidth}
    \includegraphics[width=\linewidth]{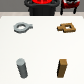}
    \caption{Nut Assembly D0}
  \end{subfigure}
  
  \caption{\textbf{8 Tasks in MimicGen Benchmark.}}
  \label{fig:mimicgen sim}
\end{figure*}

We select 8 representative tasks from MimicGen\cite{mandlekar2023mimicgen} which contains an automatically synthesizing large-scale dataset. As shown in Figure~\ref{fig:mimicgen sim}, these tasks span a wide spectrum of difficulty—as evidenced by their varying baseline success rates. Since the previous evaluations of OneDP\cite{wang2024one} and Consistency Policy\cite{kim2024consistency} utilize the absolute action space in RoboMimic, we adopt the complementary relative action representation in this part of the experiments to evaluate the adaptability of our method to different action parameterizations.

\subsubsection{Hyperparameters}
\label{sec:hyper}
We adopt the 1D-Unet backbone and ResNet-18 observation encoder architecture from Diffusion Policy~\cite{chi2023diffusion} for all policies. In the implementation of our L1 regression policy, we remove the timestep component: the U-Net takes only the observation embedding as the conditioning input, rather than concatenating observation embeddings with timestep embeddings as in other timestep-dependent policies. Apart from the components intrinsic to each algorithm, all remaining hyperparameters are kept consistent. The details of the key hyperparameters for L1 Flow training are summarized in Table \ref{tab:hyperparameters}.

\subsection{Robomimic}
\subsubsection{Simulation Setup}
RoboMimic\cite{mandlekar2021matters} is a large-scale robotic manipulation benchmark designed to study imitation learning and offline reinforcement learning. The dataset includes 5 distinct manipulation tasks, each with a dataset of demonstrations teleoperated by proficient humans. These tasks are designed to enhance the learning effectiveness of robots through real human demonstrations.

\begin{figure*}[t]
  \centering
  \begin{subfigure}{0.2\linewidth}
    \includegraphics[width=\linewidth]{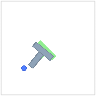}
    \caption{Push T}
  \end{subfigure}
  \hfill
  \begin{subfigure}{0.2\linewidth}
    \includegraphics[width=\linewidth]{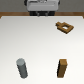}
    \caption{Square}
  \end{subfigure}
  \hfill
  \begin{subfigure}{0.2\linewidth}
    \includegraphics[width=\linewidth]{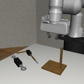}
    \caption{ToolHang}
  \end{subfigure}
  \hfill
  \begin{subfigure}{0.2\linewidth}
    \includegraphics[width=\linewidth]{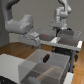}
    \caption{Transport}
  \end{subfigure}\hfill
  
  \caption{\textbf{The PushT Task and 3 Tasks in Robomimic Benchmark.}}
  \label{fig:robomimic sim}
\end{figure*}

For the experiments in Section~\ref{sec:comparison}, we choose three representative tasks from RoboMimic—\textbf{Square}, \textbf{ToolHang}, and \textbf{Transport}—as well as the \textbf{PushT} task from IBC~\cite{florence2022implicit}. Visualizations of these tasks in simulation are provided in Figure~\ref{fig:robomimic sim}.

\subsubsection{Hyperparameters}
Following the implementation in Section~\ref{sec:hyper}, we adopt the identical backbone in Diffusion Policy~\cite{chi2023diffusion}. All optimization hyperparameters are kept identical to the official implementation, except for a few task-specific settings. This ensures that any observed performance differences are attributable solely to the training objective, rather than architectural or optimization variations. Key hyperparameters and task-specific training configurations are summarized in Tables~\ref{tab:hyperparameters} and~\ref{tab:training_details}, respectively.

For certain tasks, unstable training dynamics were observed under the default learning rate ($1 \times 10^{-4}$). To promote convergence, the learning rate was reduced where needed, and the final values are reported in Table~\ref{tab:training_details}. Higher maximum epoch limits are adopted specifically for \textbf{Square-mh} (1000 epochs) and \textbf{ToolHang-ph} (500 epochs) to ensure a well-shaped cosine annealing schedule and prevent premature decay of the learning rate to near-zero. In practice, early stopping is triggered at 200 epochs for both tasks, as validation performance typically saturates by then. All models are trained on NVIDIA RTX 4090 GPUs with mixed-precision training enabled, and the corresponding wall-clock training times are reported in Table~\ref{tab:training_details}.

\begin{table}[!htbp]
\centering
\caption{\textbf{Training configurations across tasks.} Reported values include the learning rate, training epoch, and wall-clock training time (trained on a single NVIDIA RTX 4090 GPU with mixed-precision training). For \textbf{Square-mh} and \textbf{ToolHang-ph}, early stopping is triggered at epoch 200 despite higher maximum epoch settings mentioned before.}
\label{tab:training_details}
\begin{tabular}{lccc}
\toprule
\textbf{Task} & \textbf{LR} & \textbf{Epochs} & \textbf{Time (h)} \\
\midrule
PushT       & 1e-4 & 200 & 2.5  \\
Square-mh   & 2e-5 & 200 & 18  \\
Square-ph   & 1e-4 & 200 & 10  \\
ToolHang-ph & 5e-5 & 200 & 53  \\
Transport-ph& 6e-5 & 200 & 76  \\
\bottomrule
\end{tabular}
\end{table}

\subsection{Real-World Experiment}
We provide the visualization of the policy rollouts with execution time stamp to clarify the efficiency of the proposed method in real-world, as shown in Figure \ref{fig:real-world}. L1Flow achieves substantially faster inference which is roughly 10\(\times\) speed-up over DDIM (16 steps) and about 70\(\times\) over DDPM (100 steps). However, due to hardware limitations, the actual end-to-end completion time yields at most a ~3\(\times\) speed-up. We also observe that, at the same spatial locations, DDIM is more prone to producing confusing actions compared with DDPM and L1Flow. This often manifests as the policy getting ``stuck'' momentarily, which in turn leads to noticeably longer task completion times.

\begin{table*}[ht]
\centering
\caption{\textbf{Key hyperparameters for L1 Flow training.}}
\label{tab:hyperparameters}
\begin{tabular}{lc}
\toprule
\textbf{Hyperparameters} & \textbf{Values} \\
\midrule
optimizer & AdamW($\beta_1=0.95, \beta_2=0.999$, weight decay=$1\times10^{-6}$) \\
learning rate & 1e-4 in MimicGen, see Table \ref{tab:training_details} for MimicGen\\
learning rate scheduler & cosine annealing schedule with 500 warmup steps \\
batch size & 128 in MimicGen, 64 in Robomimic\&PushT \\
training epochs & 500 in MimicGen, 200 in Robomimic\&PushT\\
action chunk size (horizon) & 16 \\
executed actions per step & 8 \\
observed steps per decision & 2 \\
EMAs used & Yes ($\gamma=1.0$, power=0.75) \\
abs action used & No in MimicGen, Yes in Robomimic\&PushT \\
\bottomrule
\end{tabular}
\end{table*}

\begin{figure*}[ht]
  \centering
  \begin{subfigure}{0.24\linewidth}
    \includegraphics[width=\linewidth]{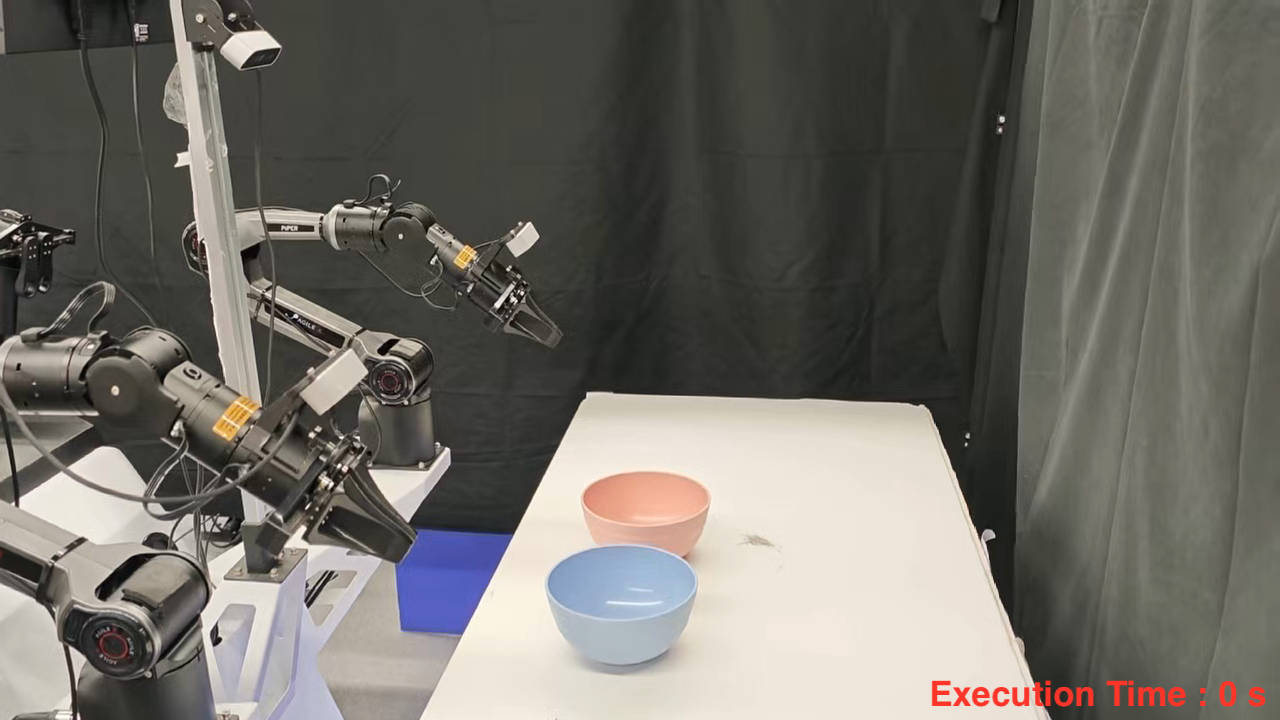}
  \end{subfigure}
  \begin{subfigure}{0.24\linewidth}
    \includegraphics[width=\linewidth]{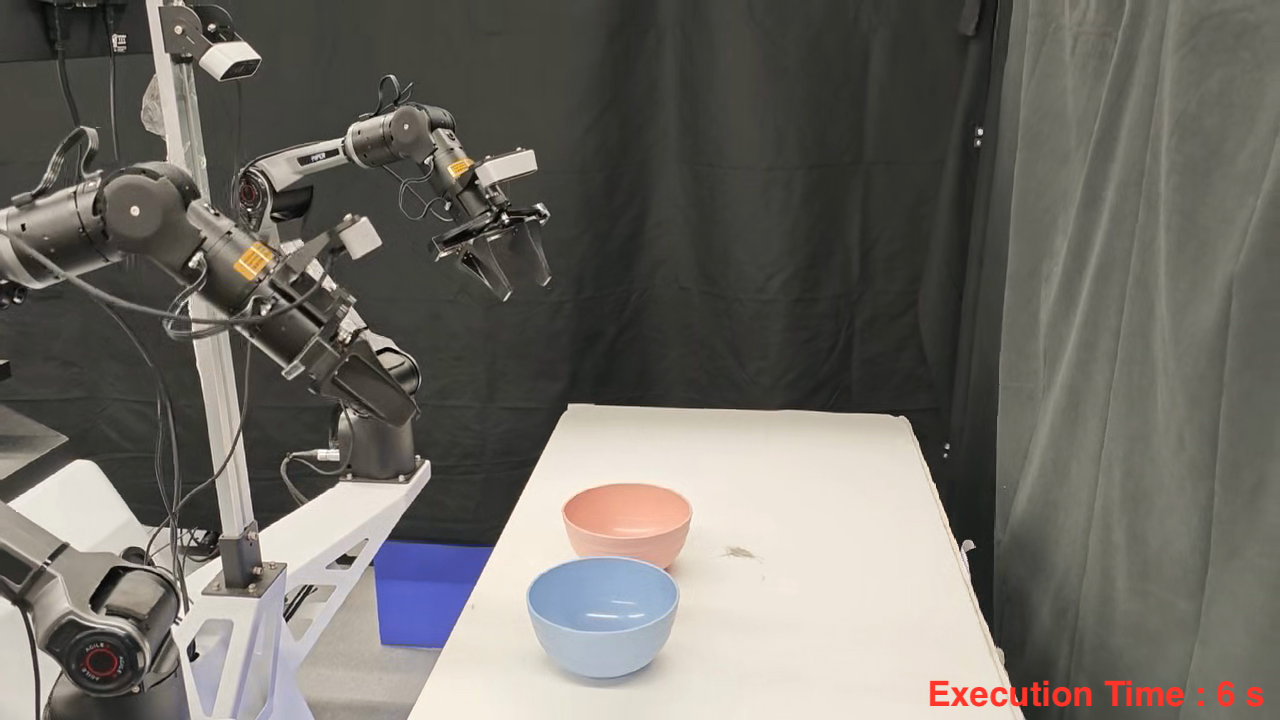}
  \end{subfigure}
  \begin{subfigure}{0.24\linewidth}
    \includegraphics[width=\linewidth]{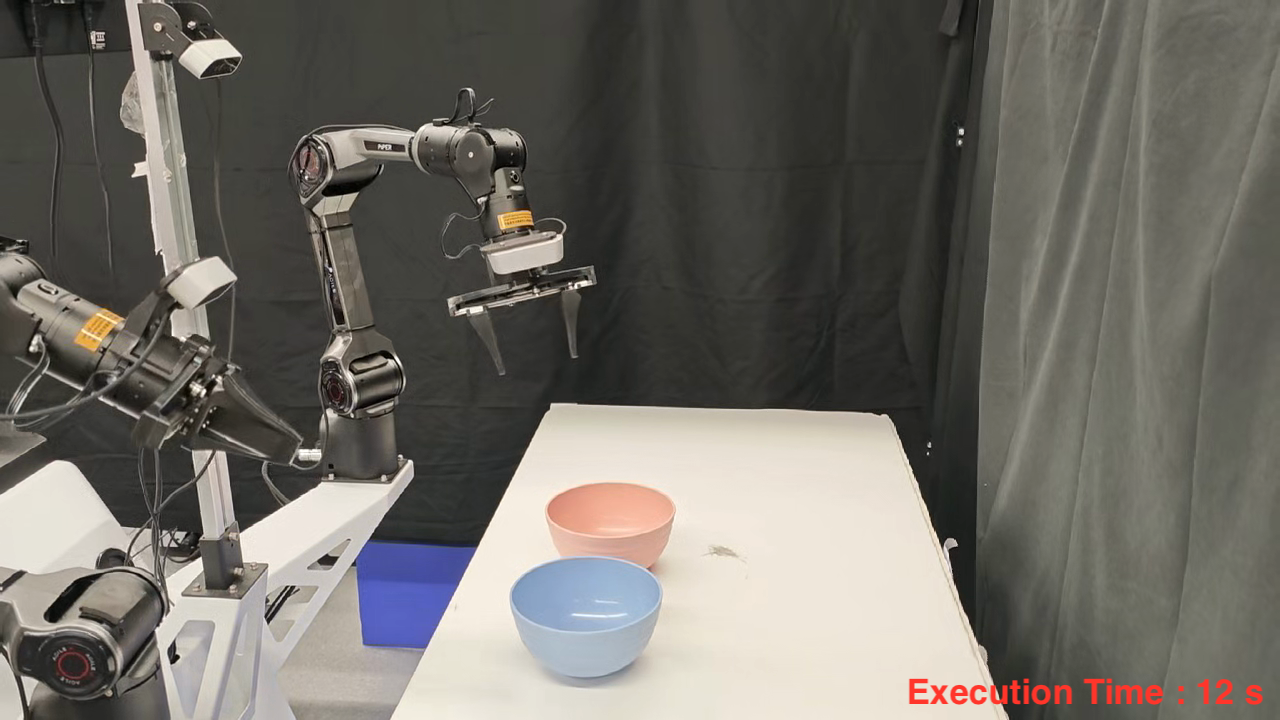}
  \end{subfigure}
  \begin{subfigure}{0.24\linewidth}
    \includegraphics[width=\linewidth]{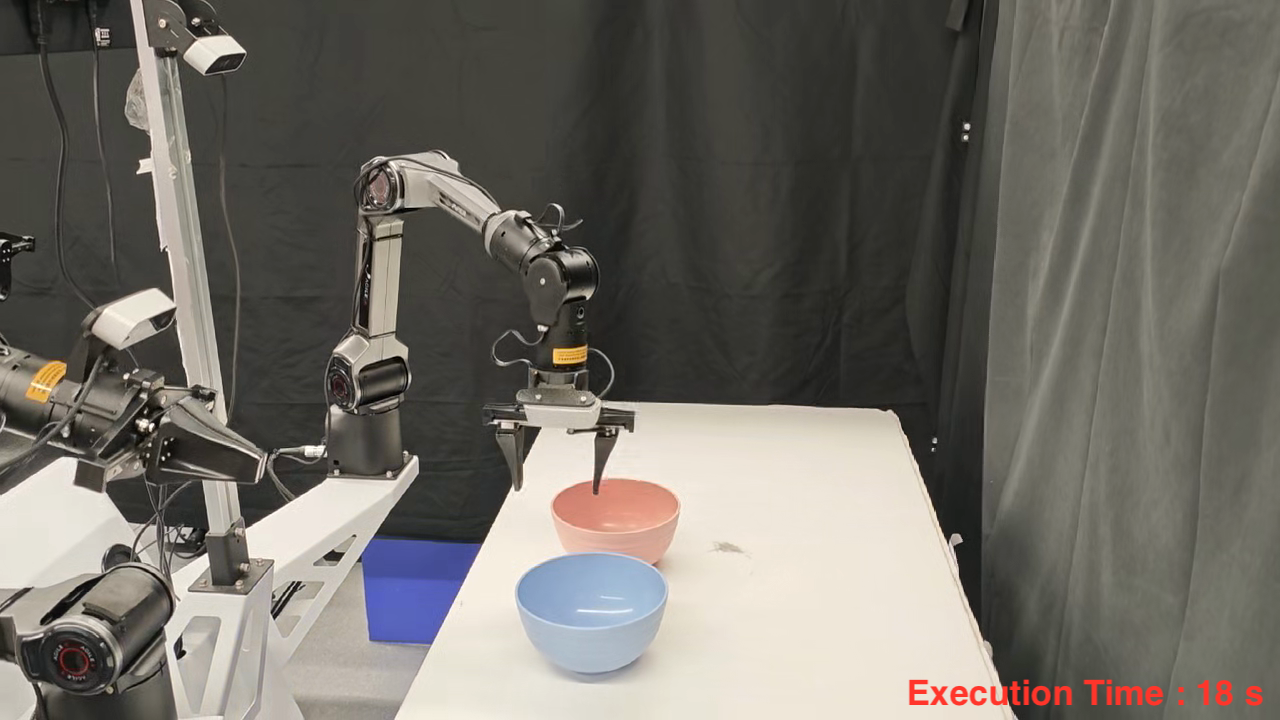}
  \end{subfigure}
  \begin{subfigure}{0.24\linewidth}
    \includegraphics[width=\linewidth]{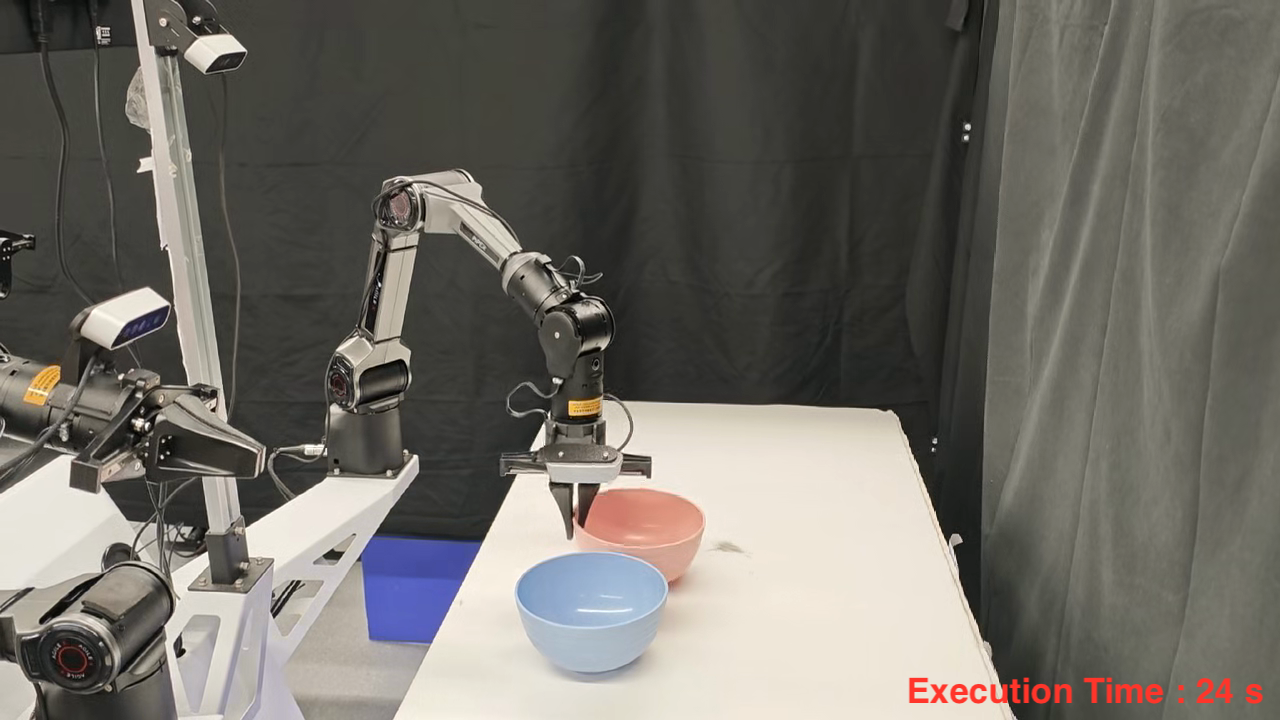}
  \end{subfigure}
  \begin{subfigure}{0.24\linewidth}
    \includegraphics[width=\linewidth]{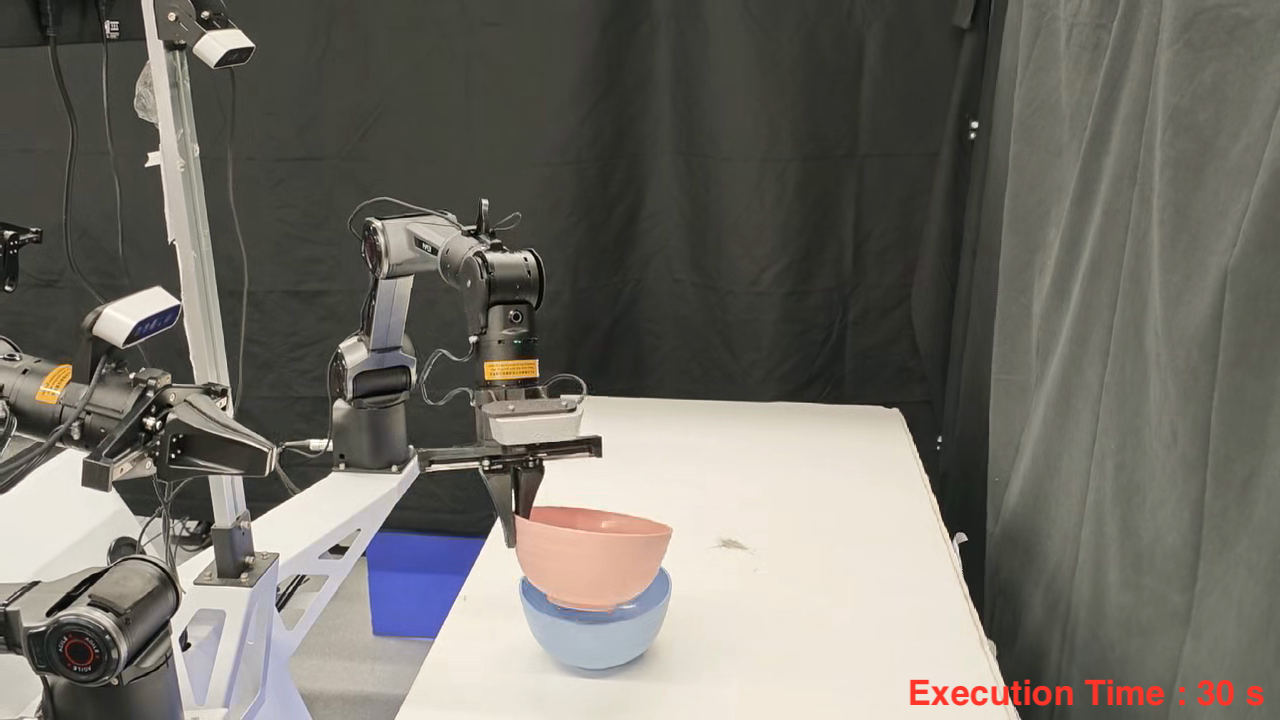}
  \end{subfigure}
  \begin{subfigure}{0.24\linewidth}
    \includegraphics[width=\linewidth]{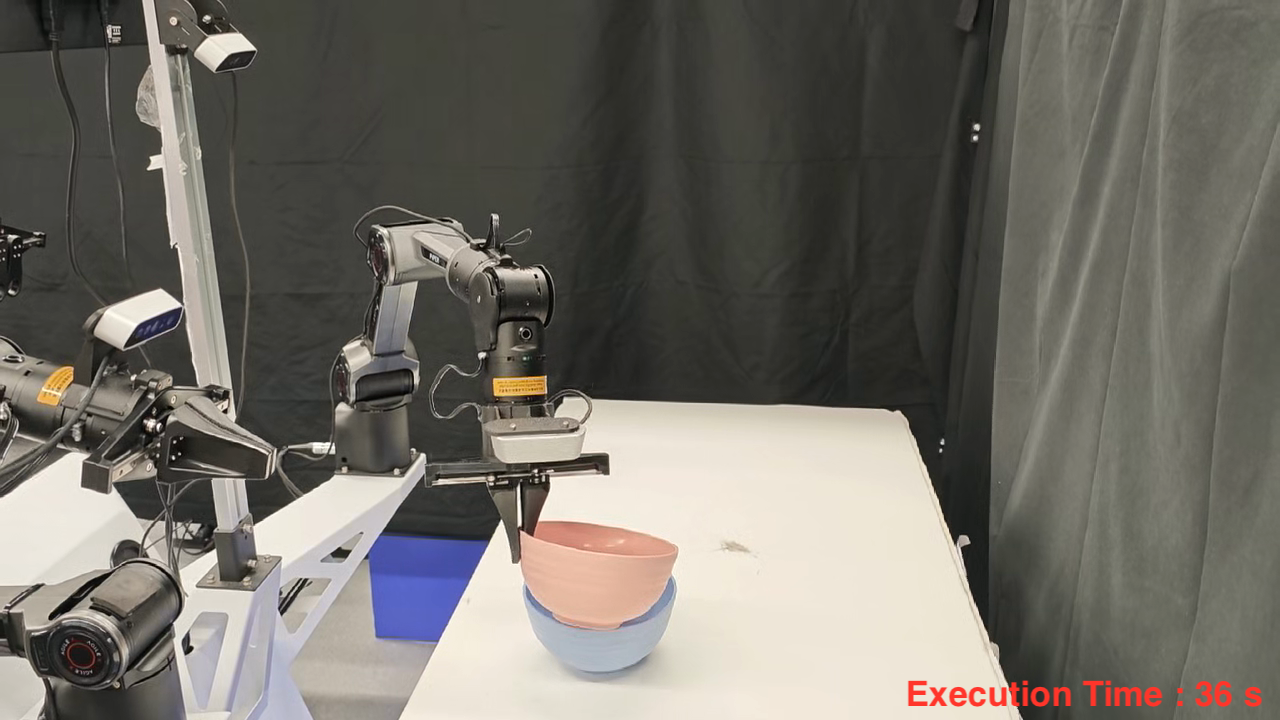}
  \end{subfigure}
  \begin{subfigure}{0.24\linewidth}
    \includegraphics[width=\linewidth]{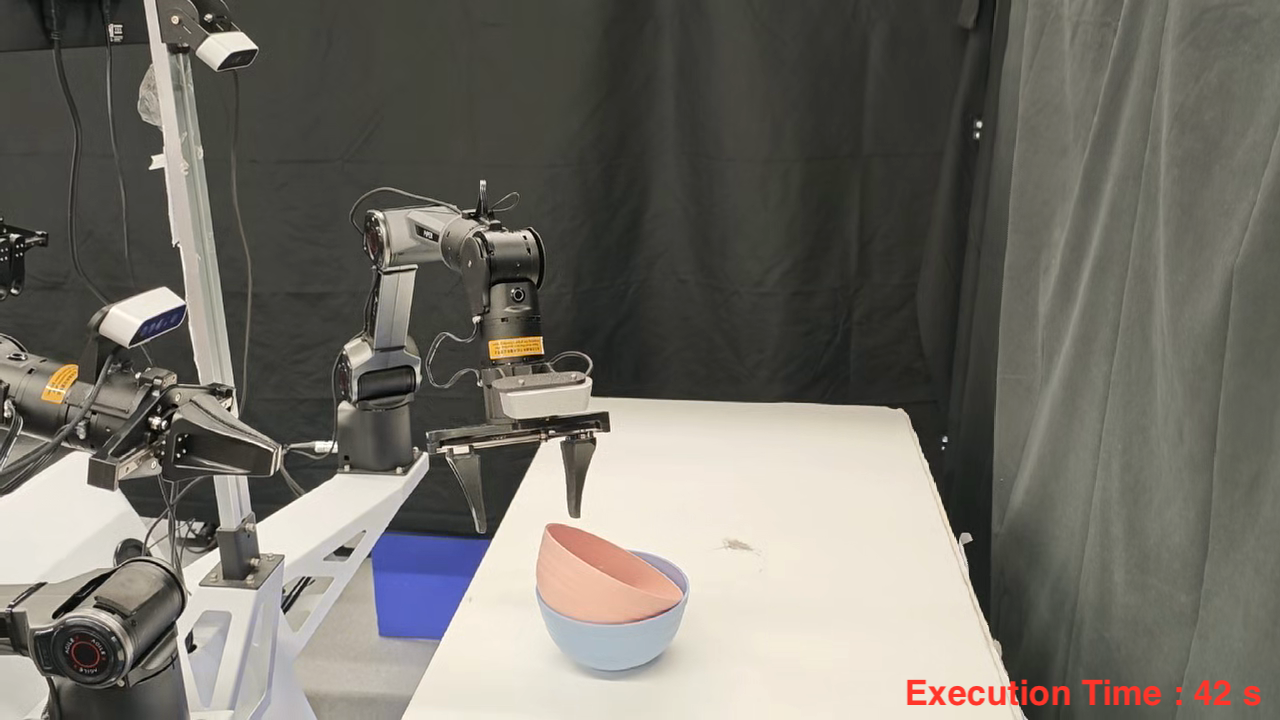}
  \end{subfigure}
  \caption{\textbf{A Real Execution Example of DDPM (100).} The total completion time is around 42 seconds. Each action chunk requires roughly 1 second for DDPM inference, leading to noticeable execution lag and prolonged task completion time.}
  \label{fig:real ddpm}
\end{figure*}

\begin{figure*}[ht]
  \centering
  \begin{subfigure}{0.24\linewidth}
    \includegraphics[width=\linewidth]{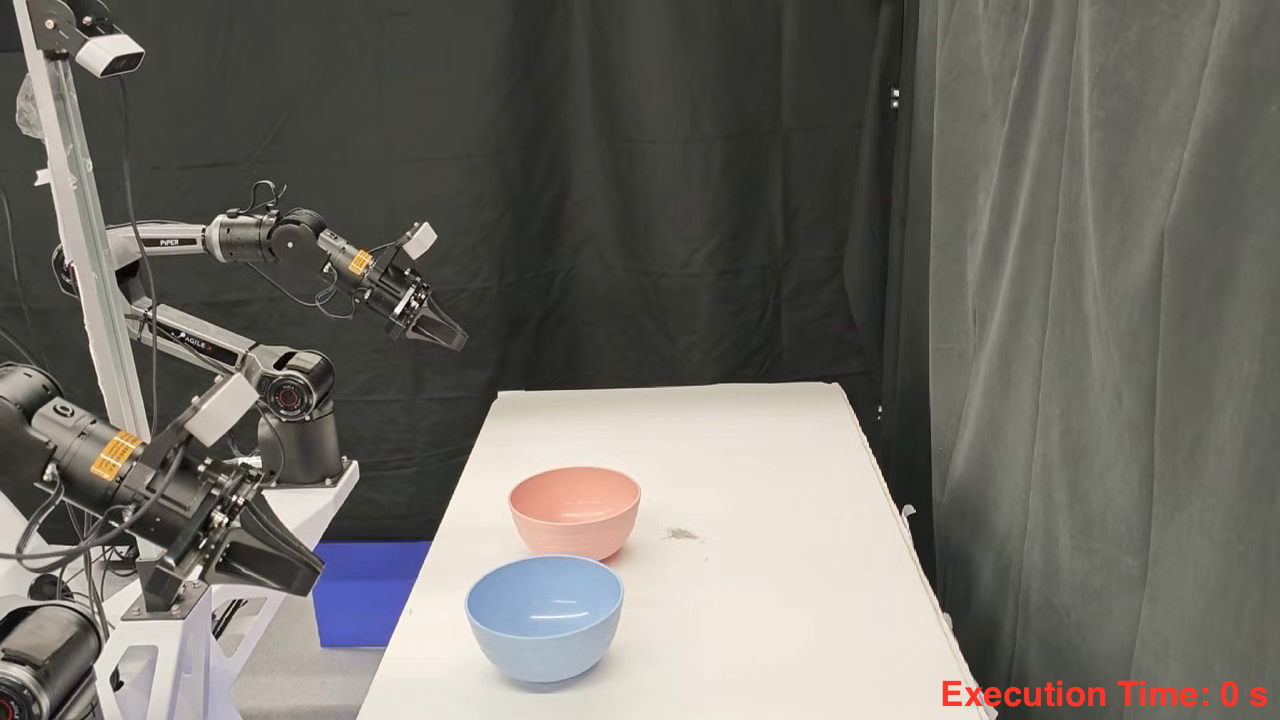}
  \end{subfigure}
  \begin{subfigure}{0.24\linewidth}
    \includegraphics[width=\linewidth]{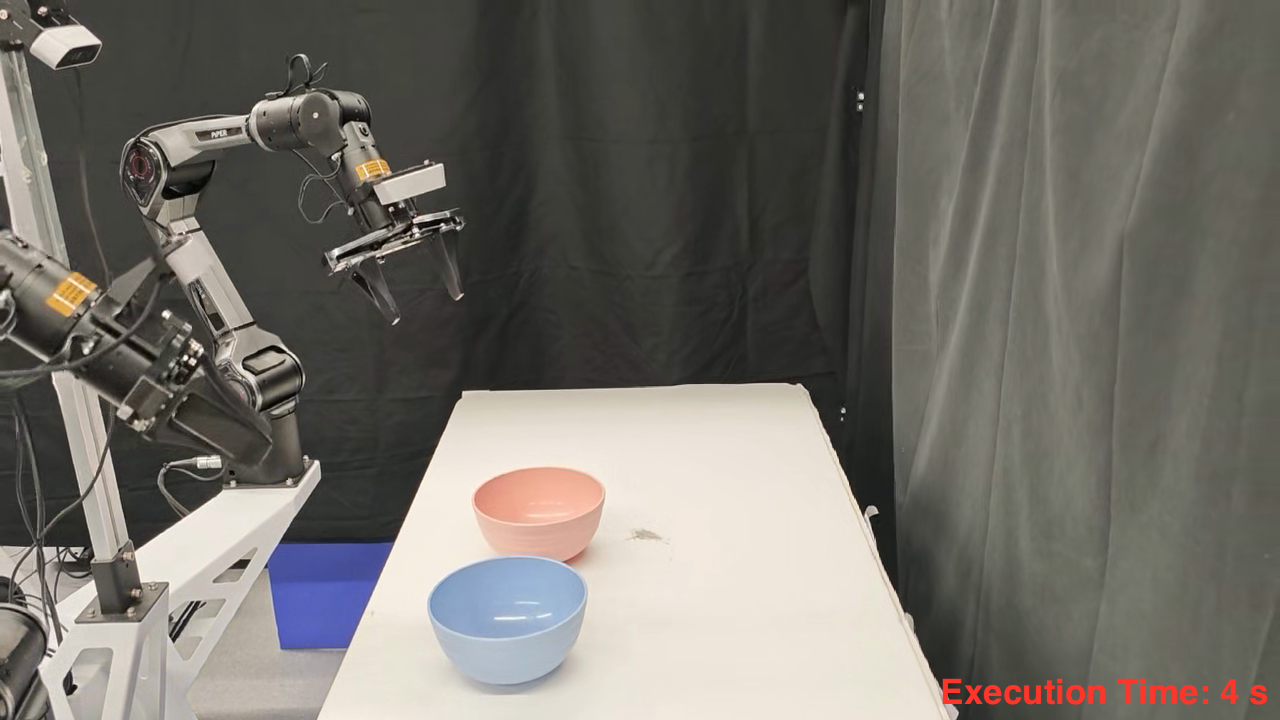}
  \end{subfigure}
  \begin{subfigure}{0.24\linewidth}
    \includegraphics[width=\linewidth]{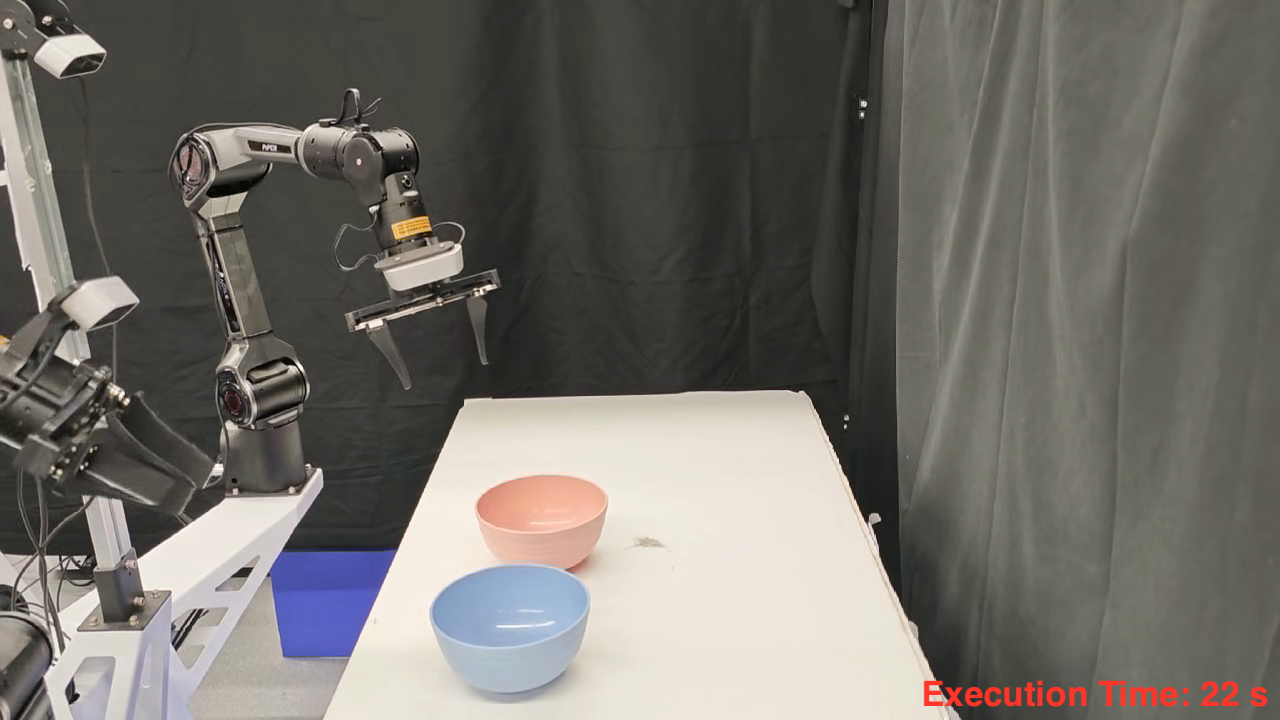}
  \end{subfigure}
  \begin{subfigure}{0.24\linewidth}
    \includegraphics[width=\linewidth]{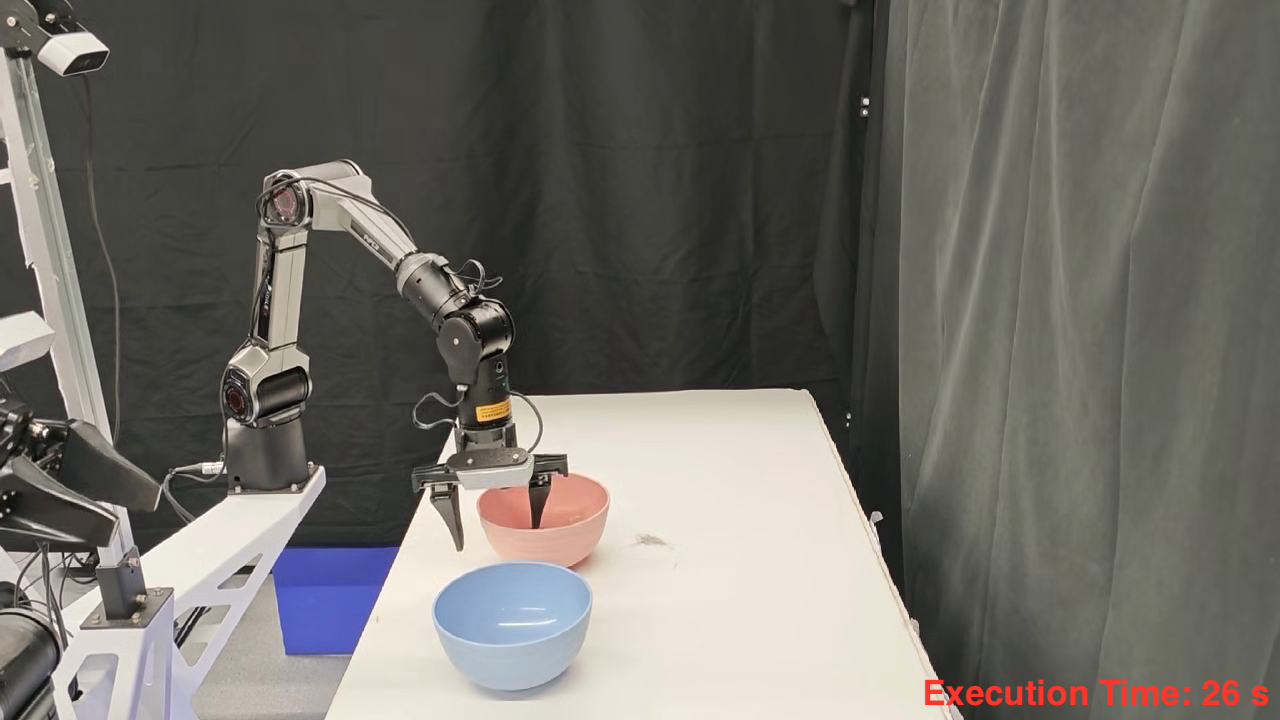}
  \end{subfigure}
  \begin{subfigure}{0.24\linewidth}
    \includegraphics[width=\linewidth]{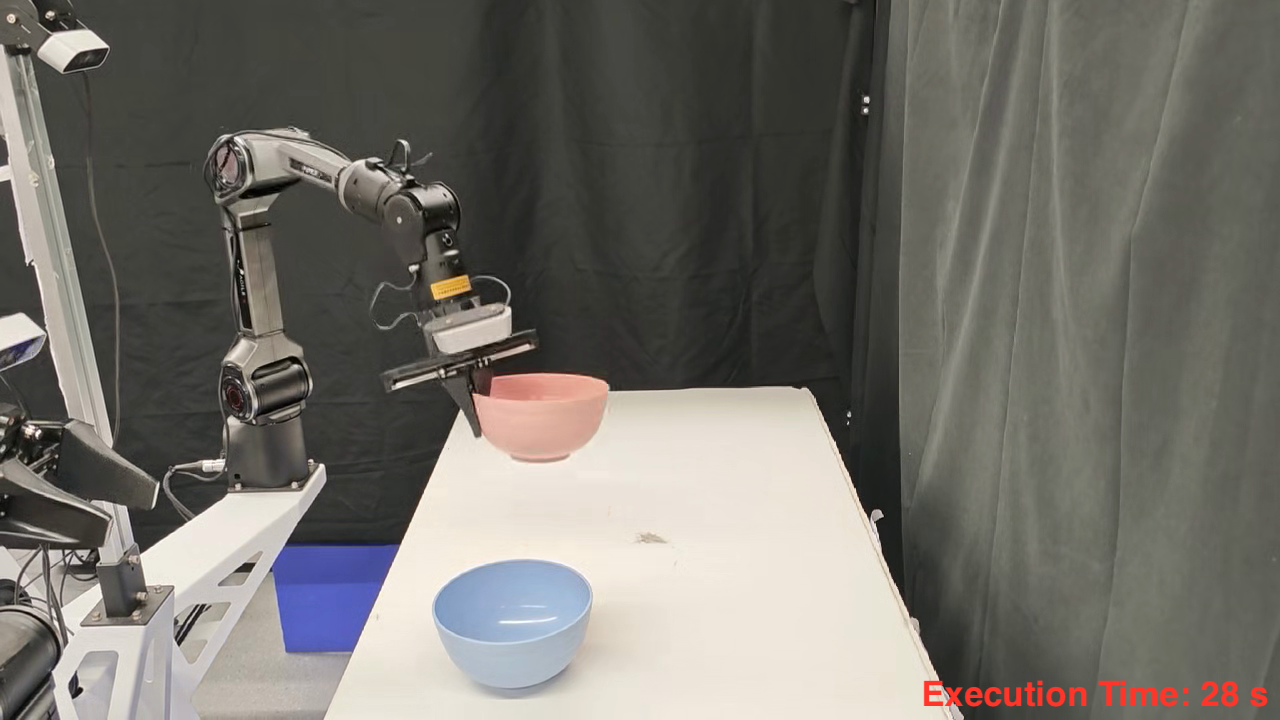}
  \end{subfigure}
  \begin{subfigure}{0.24\linewidth}
    \includegraphics[width=\linewidth]{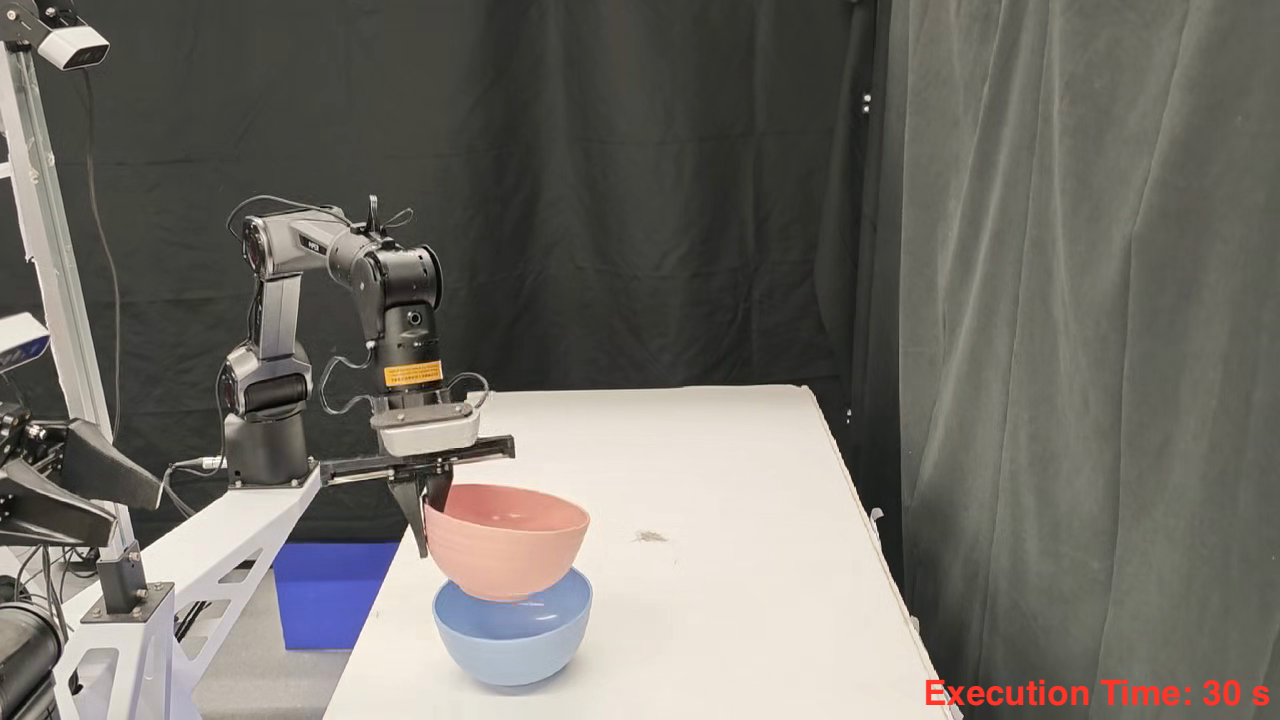}
  \end{subfigure}
  \begin{subfigure}{0.24\linewidth}
    \includegraphics[width=\linewidth]{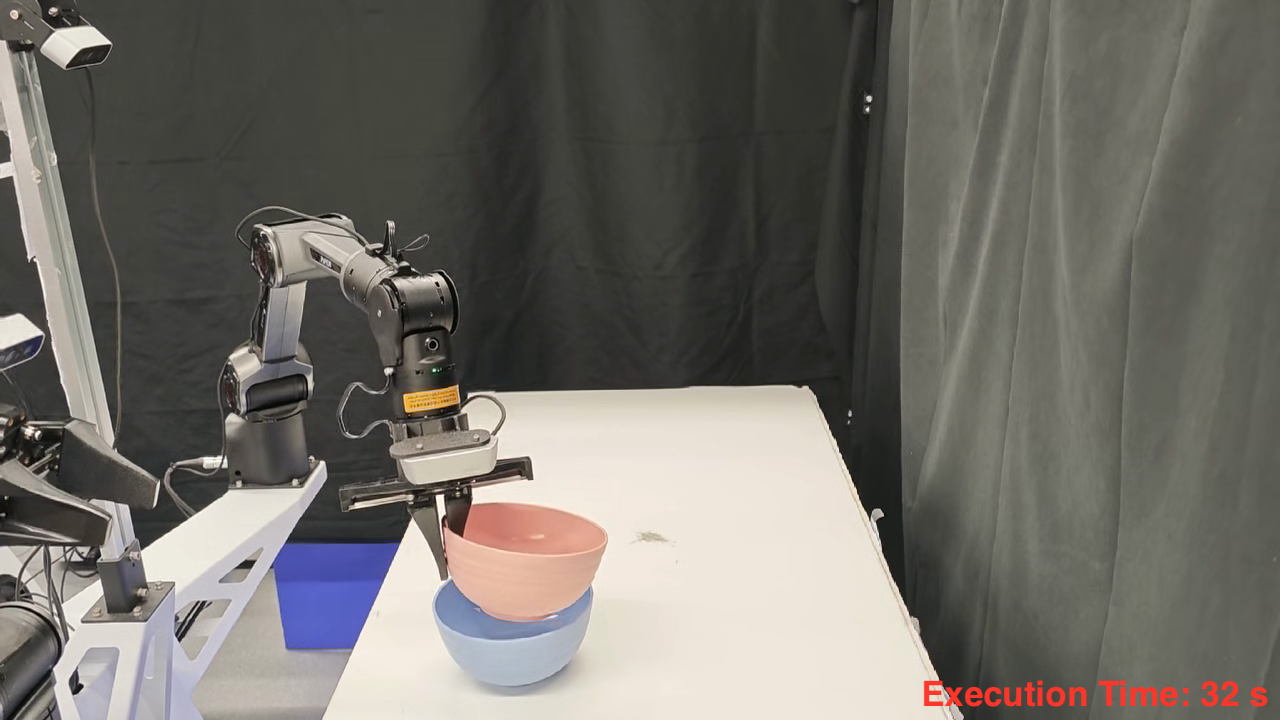}
  \end{subfigure}
  \begin{subfigure}{0.24\linewidth}
    \includegraphics[width=\linewidth]{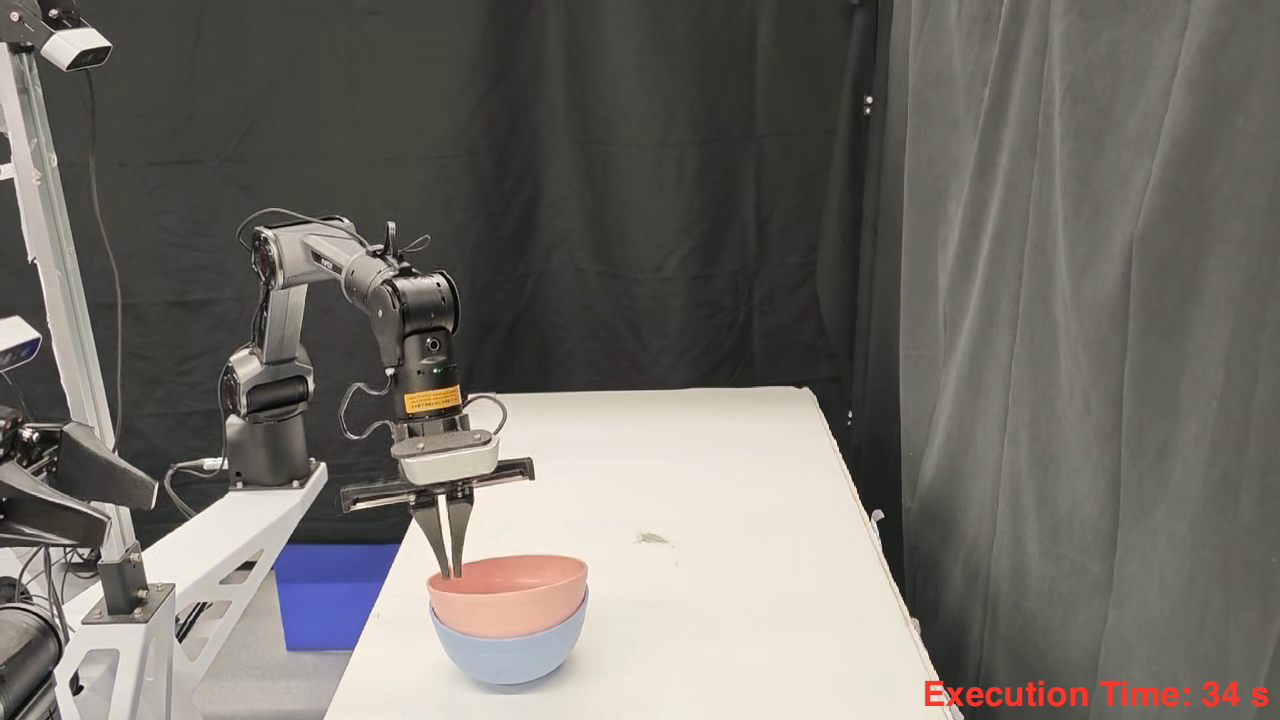}
  \end{subfigure}
  \caption{\textbf{A Real Execution Example of DDIM (16).} The total completion time is around 36 seconds. The DDIM policy is occasionally stuck during the grasping and placing stage which leads to longer completion time. }
  \label{fig:real ddim}
\end{figure*}

\begin{figure*}[ht]
  \centering
  \begin{subfigure}{0.24\linewidth}
    \includegraphics[width=\linewidth]{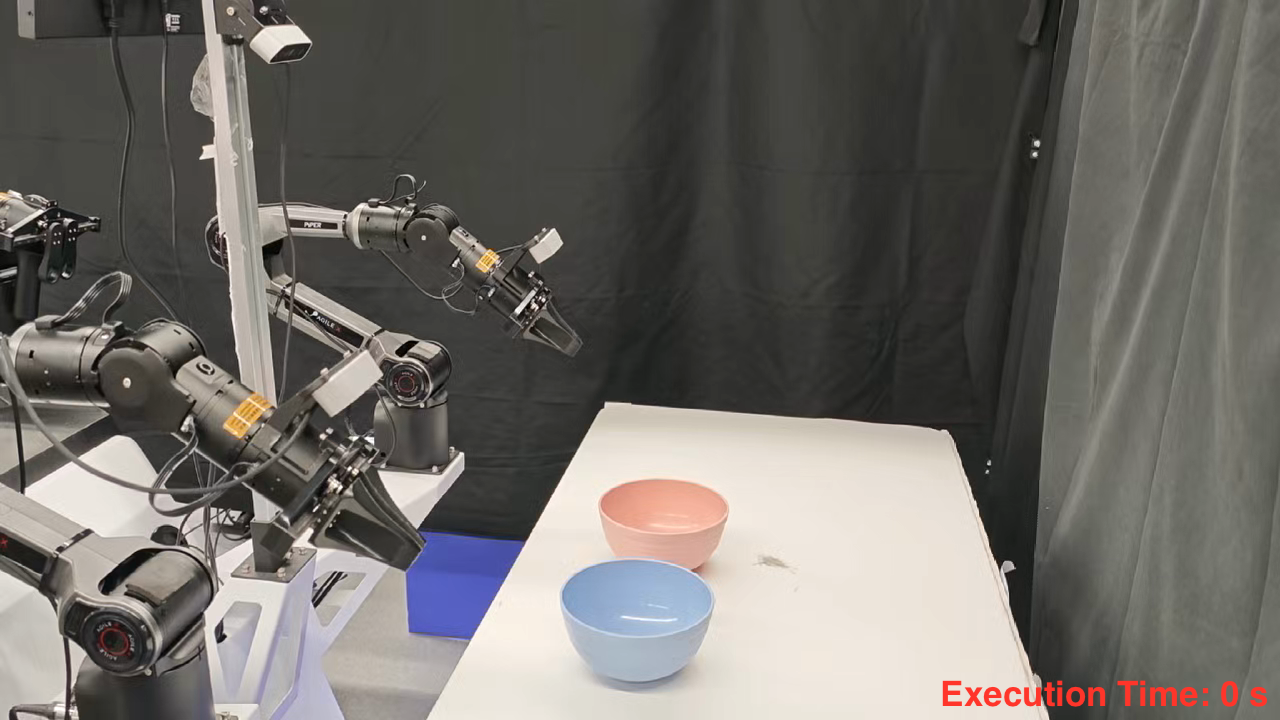}
  \end{subfigure}
  \begin{subfigure}{0.24\linewidth}
    \includegraphics[width=\linewidth]{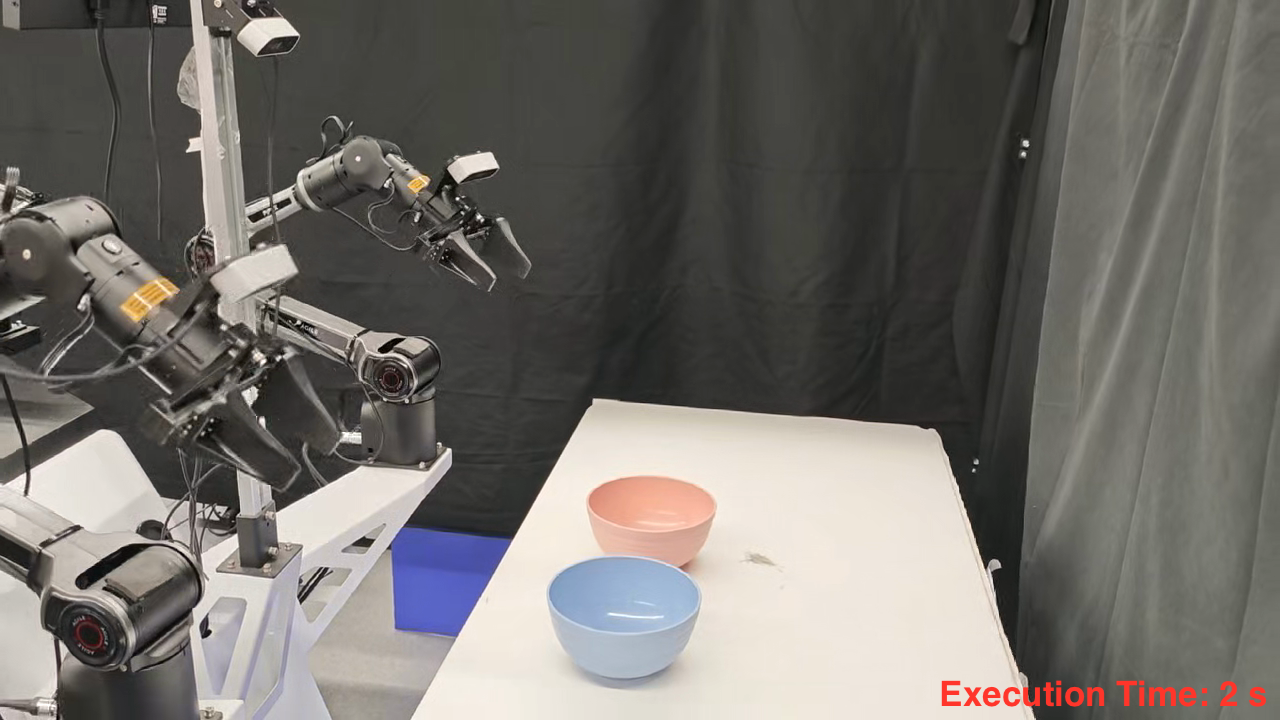}
  \end{subfigure}
  \begin{subfigure}{0.24\linewidth}
    \includegraphics[width=\linewidth]{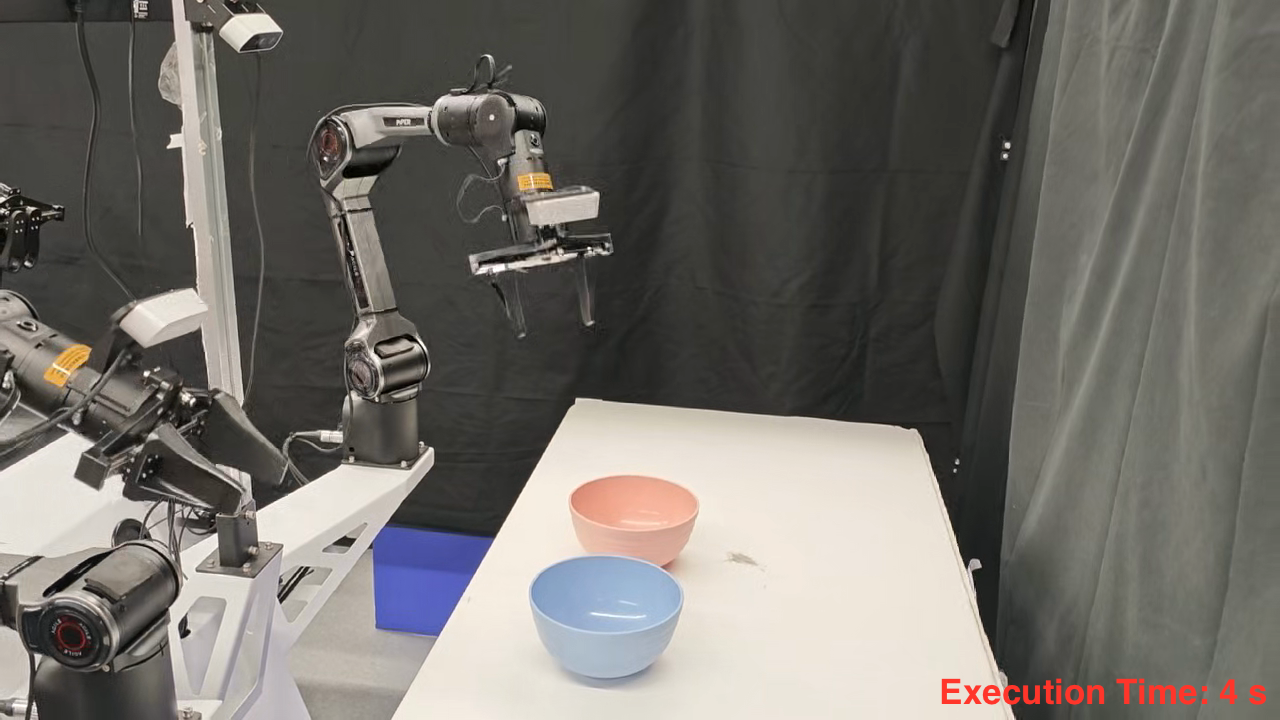}
  \end{subfigure}
  \begin{subfigure}{0.24\linewidth}
    \includegraphics[width=\linewidth]{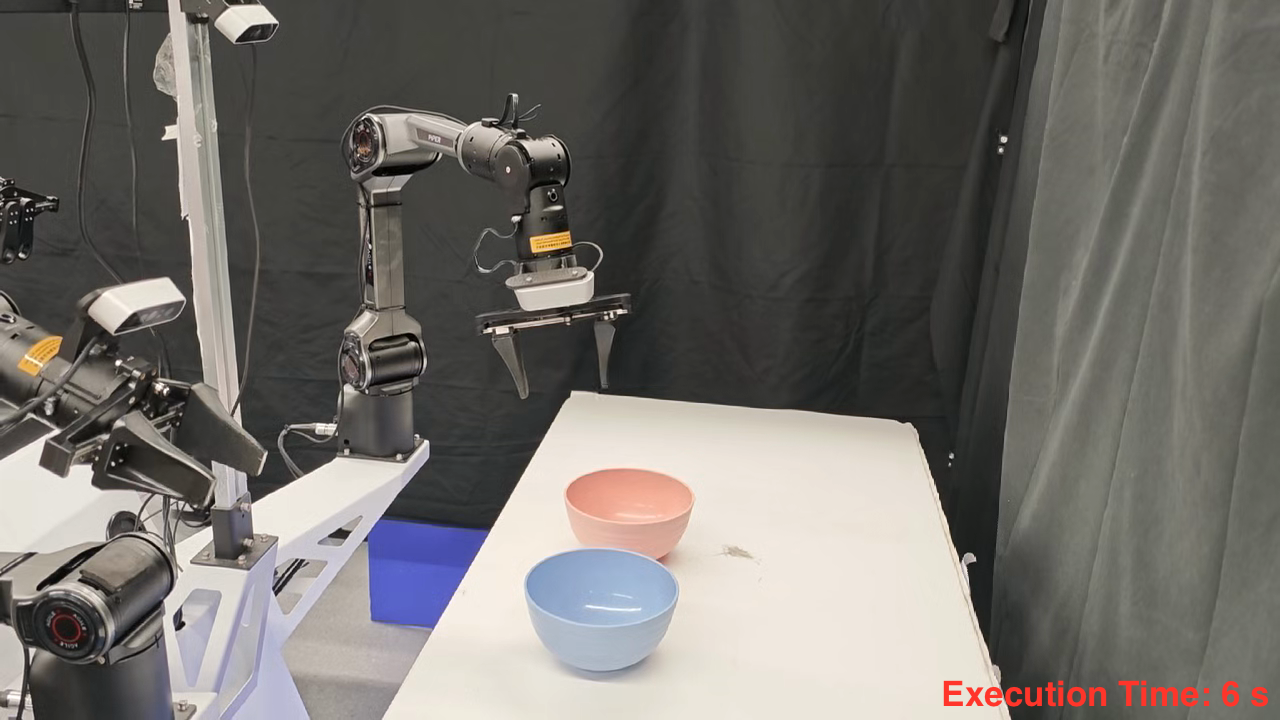}
  \end{subfigure}
  \begin{subfigure}{0.24\linewidth}
    \includegraphics[width=\linewidth]{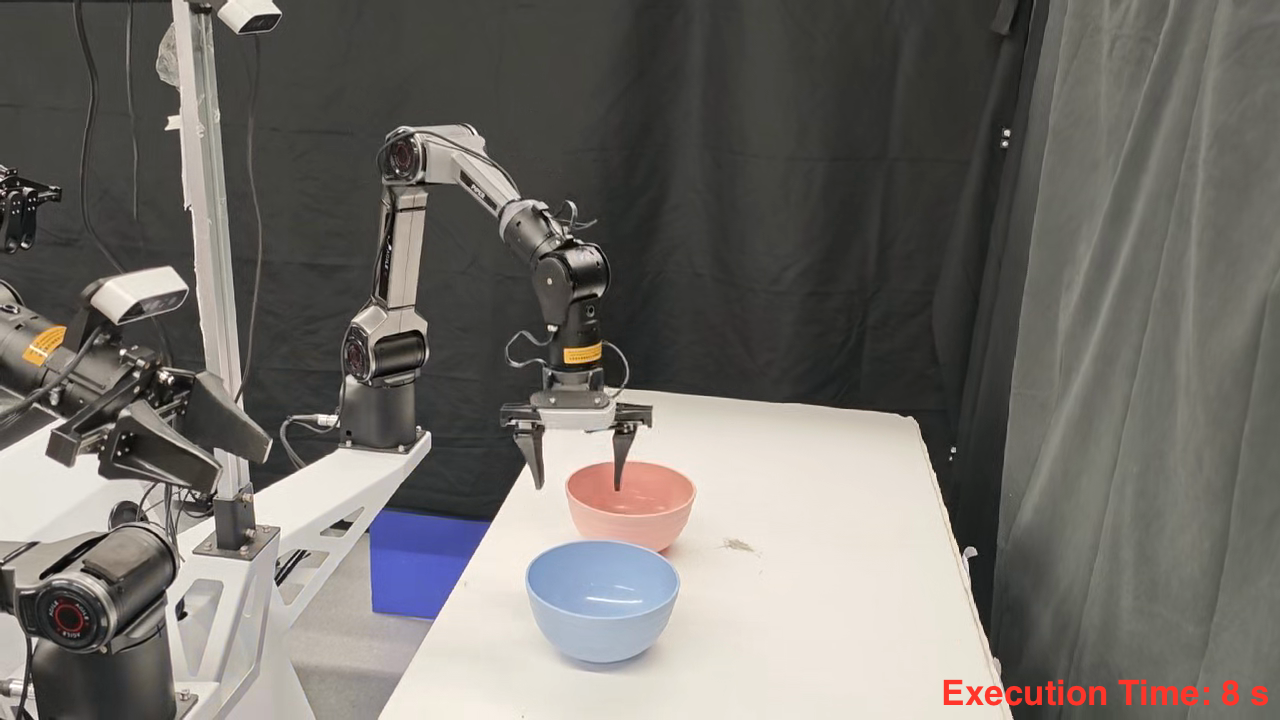}
  \end{subfigure}
  \begin{subfigure}{0.24\linewidth}
    \includegraphics[width=\linewidth]{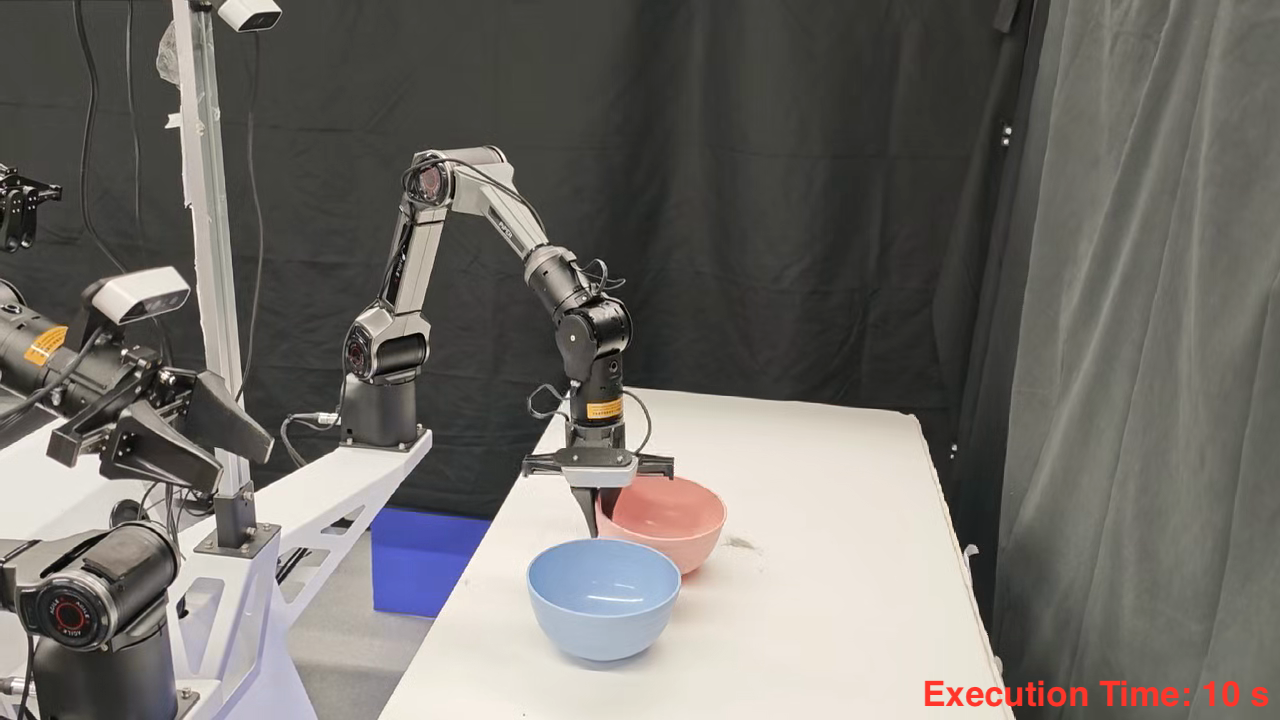}
  \end{subfigure}
  \begin{subfigure}{0.24\linewidth}
    \includegraphics[width=\linewidth]{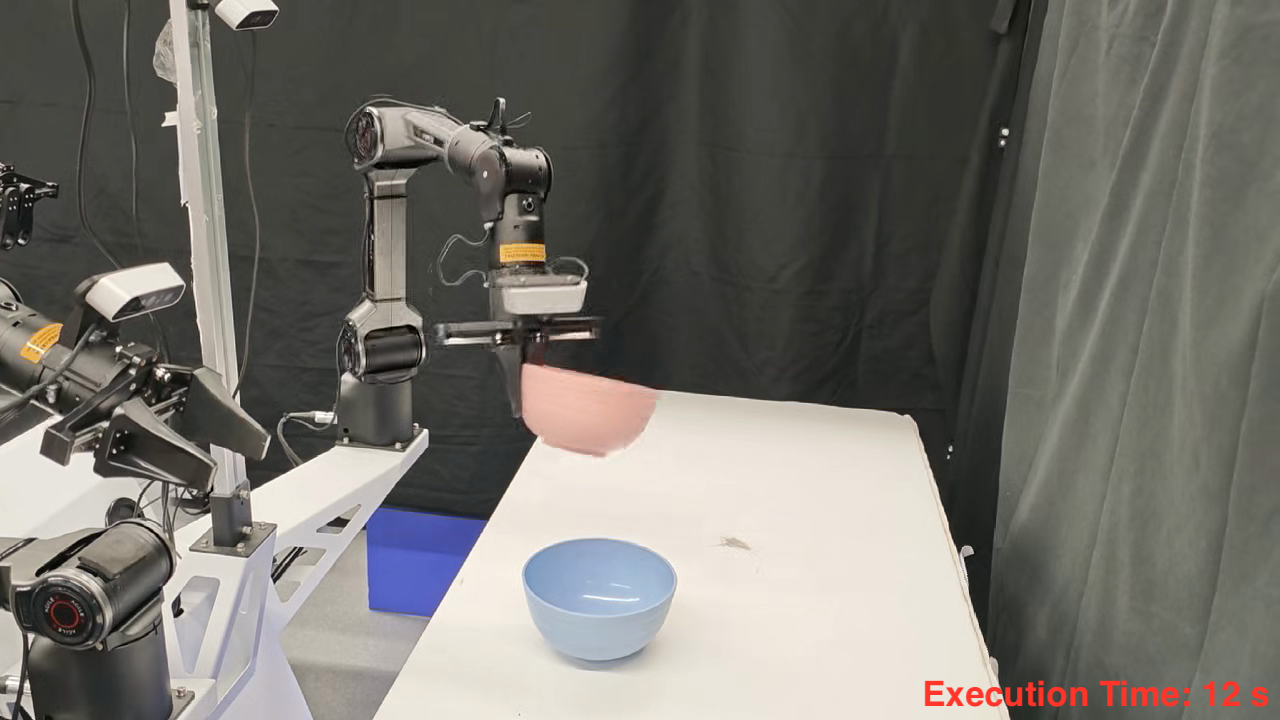}
  \end{subfigure}
  \begin{subfigure}{0.24\linewidth}
    \includegraphics[width=\linewidth]{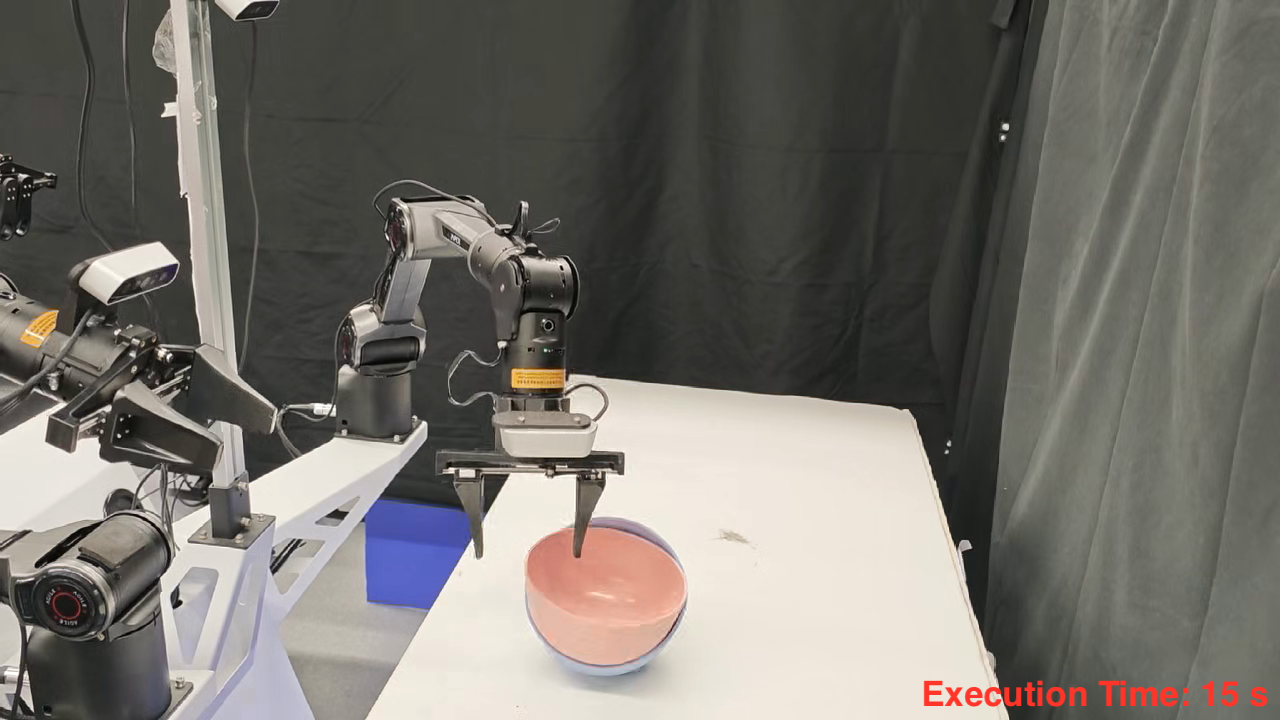}
  \end{subfigure}
  \caption{\textbf{A Real Execution Example of L1 Flow (Ours).} The total completion time is around 15 seconds. Due to hardware limitations, the actual end-to-end completion time yields at most a ~3\(\times\) speed-up.}
  \label{fig:real l1flow}
\end{figure*}

\section{Code Example}
We provide a minimal implementation of \textbf{L1 Flow}, as shown in the Figure \ref{fig:code}, which can be directly substituted for the corresponding components in standard diffusion or flow-matching pipelines.

\begin{figure*}
\centering
\begin{lstlisting}[language=Python,caption={Training}]
noise = torch.randn((batchsize,horizon,action_dim), device=device)

# logistic+uniform
timesteps = logisticnormal_dist.sample((batchsize,))[:,0].to(device)
uni_timesteps = torch.rand_like(timesteps)
mask = torch.rand_like(uni_timesteps) < 0.01
timesteps[mask] = uni_timesteps[mask]

noisy_action = timesteps * action + (1 - timesteps) * noise

# Predict the sample
x_pred = self.model(noisy_action, timesteps, cond=cond)

# Compute the loss
loss = F.l1_loss(x_pred, action)
\end{lstlisting}

\begin{lstlisting}[language=Python,caption={Inference}]
action = torch.randn(
            size=((batchsize, horizon, action_dim), 
            device=device)
    
# set step values
dt = 0.5
t = torch.zeros(1, device=device)
        
# one-step integration
x_pred = model(action, t, 
            cond=cond)
v_t = (x_pred - action)/(1-t)
action = action + dt*v_t
t = t + dt
        
# direct prediction
x_pred = model(action, t, cond=cond) 
\end{lstlisting}
\caption{L1 Flow Code Example.}
\label{fig:code}
\end{figure*}


\end{document}